\pdfoutput=1

\documentclass[11pt]{article}

\usepackage[final]{acl}

\usepackage{times}
\usepackage{latexsym}

\usepackage[T1]{fontenc}

\usepackage[utf8]{inputenc}

\usepackage{microtype}
\usepackage{svg}
\usepackage{amsmath}

\usepackage{tabularx}
\usepackage{multirow}

\usepackage[utf8]{inputenc} 
\usepackage[T1]{fontenc}    
\usepackage{hyperref}       
\usepackage{url}            
\usepackage{booktabs}       
\usepackage{amsfonts}       
\usepackage{nicefrac}       
\usepackage{microtype}      
\usepackage{xcolor}         

\usepackage{subcaption}

\usepackage{ntheorem}

\usepackage[nameinlink]{cleveref}
\usepackage[shortlabels]{enumitem}

\title{Linear Interpolation In Parameter Space is Good Enough for Fine-Tuned Language Models}

\author{Mark Rofin, Nikita Balagansky, and Daniil Gavrilov
       \\
       Tinkoff\\
           \{ext.mrofin, n.n.balaganskiy, d.gavrilov\}@tinkoff.ai
       }

\begin{document}

\maketitle

\begin{abstract}
The simplest way to obtain continuous interpolation between two points in high dimensional space is to draw a line between them\footnote{That applies if you are operating in a Euclidean space.}.
While previous works focused on the general connectivity between model parameters, we explored linear interpolation for parameters of pre-trained models after fine-tuning.  Surprisingly, we could perform linear interpolation without a performance drop in intermediate points for fine-tuned models.
For controllable text generation, such interpolation could be seen as moving a model towards or against the desired text attribute (e.g., positive sentiment), which could be used as grounds for further methods for controllable text generation without inference speed overhead.
\end{abstract}

\section{Introduction}

Currently, large pre-trained transformer models can be considered a default choice for various NLP tasks. Training these models is a complex non-linear task that is usually performed by feeding the model a large training corpus and training it in a self-supervised manner \citep{bert, albert, roberta, gpt}. Weights obtained by this process are used either for standard fine-tuning or other methods that can be considered more effective in terms of trainable parameters \citep{v1, v2, prefix, google_prompt_tuning, Adapters, Lora}.




Since initialization using pre-trained parameters is crucial for the final model's performance, it is fascinating to observe the changes in parameters during the fine-tuning process on downstream tasks.

While recent works (\citealt{qualitatively_characterizing}, \citealt{analyzing_monotonic_linear_interpolation}) explored changes in parameter space during training, there is still little known about the details of this process, specifically for model fine-tuning. In our work, we are exploring properties of the fine-tuned language models obtained by linear interpolation.


\begin{figure}
    \centering
    \includegraphics[width=1\linewidth]{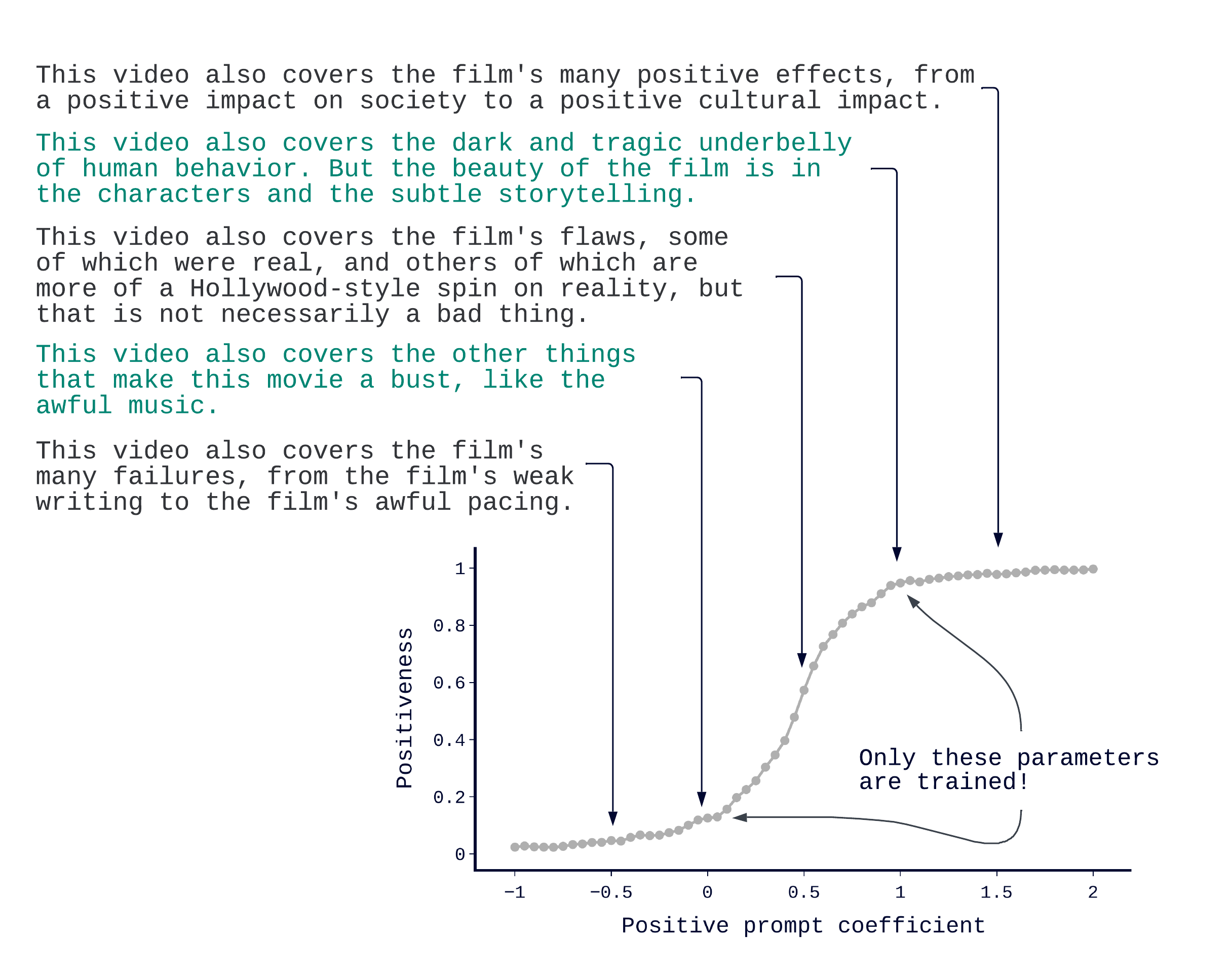}
    \caption{We experimented with linear interpolation for fine-tuned Language Models. We observed that we could fine-tune a pre-trained model on two domains (e.g., positive and negative movie reviews) and interpolate between trained weights without loss in perplexity in between these models. Furthermore, we could expand interpolation beyond trained models and get more positive or negative models than fine-tuned ones.}
    \label{fig:my_label}
\end{figure}

Surprisingly, we observed compelling evidence of the linearity of some parameter subspace of pre-trained models. The formula behind interpolation is simply 
$$\alpha \theta^- + (1 - \alpha) \theta^+,$$
where $\alpha \in \mathbb{R}$ is an interpolation weight\footnote{Note that $\alpha$ does not have to be restricted to be $\in [0; 1]$, since we found that it could exceed these boundaries during our experiments.}, and $\theta^-$, $\theta^+$ are model parameters. If both $\theta^-$ and $\theta^+$ are fine-tuned LMs (e.g., with negative and positive sentiment generations), the model parameters obtained by applying this formula are well-behaved in terms of perplexity. Therefore, the probability of positive sentiment occurring in the generated text is smoothly growing with the interpolation coefficient weight.

\begin{figure*}[h!]
    \centering
    \includegraphics[width=\linewidth]{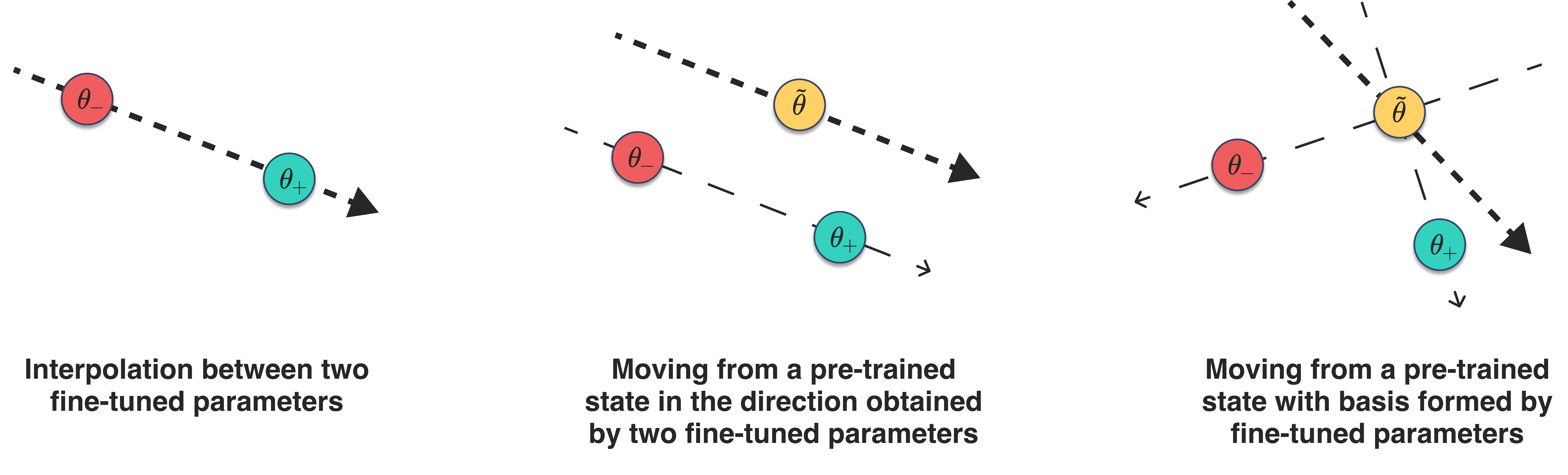}  
    \caption{Schematic overview of interpolation schemas used in our experiments. (Left) Interpolation between two fine-tuned model parameters $g_1(\alpha)$. (Center) Direction obtained by two fine-tuned model parameters used to move pre-trained model parameters to obtain $g_2(\alpha)$. (Right) Two directions formed by fine-tuned parameters and pre-trained weight define the basis used to define the direction in which the pre-trained parameter is moved to get $g_3(\alpha, \beta)$. See Section \ref{linear_interpolation_section} for more details.
}
    \label{fig:schematic}
\end{figure*}

We investigated the reasons for this phenomenon and found that the same initialization from pre-trained models is crucial for the linear properties.

Utilizing the parameter space is interesting in terms of theoretical results and insights, but what is more important is that it can be used for practical tasks. E.g., linear interpolation makes it possible to apply two attributes in a condition at the same time, or improve attribute presence with desired weight without any computational overhead.

\section{Related Work}

\citealt{qualitatively_characterizing} found that the loss landscape during interpolation between initial weights and weights after training has no significant peaks and decreases monotonically during interpolation. This is interesting since training is a complex non-linear task, and model weights tend to fall into a local optima point after training is complete. Continuing this line of research \citealt{analyzing_monotonic_linear_interpolation} found a link between batch normalization \cite{batch_normalization} and linearity of the logits' path during training.

These observations raise a question about how we can interpolate between two local optima without a loss in quality. \citealt{linear_mode_connectivity_and_lth} discovered evidence showing that finding a winning ticket \cite{lottery_ticket_hypothesis} during iterative pruning is closely connected to finding linear connectivity between optimal points in a weight space. In addition, \citealt{linear_mode_connectivity_and_lth} proposed the \textit{Loss Barrier} metric for evaluating the connectivity between parameters of two models.



\citealt{the_role_of_permutation} explored the impact of the width and depth of networks on their connectivity. Their findings showed that the wider the network is, the lower its loss barrier. Meanwhile, the deeper the network is, the higher its barrier value. Furthermore, \citealt{the_role_of_permutation} proposed a conjecture about weight permutations and solutions obtained by gradient descent. More precisely, most SGD solutions belong to a set $S$, whose elements can be permuted in such a way that there is no barrier to the linear interpolation between any two permuted elements in $S$. \citealt{git_re_basin} proposed several methods for sufficient permutation in order to reduce the loss barrier.



To further explore why a zero loss barrier is possible, the Lazy Training theory \cite{lazy_training} can be used. I.e., if a neural network has sufficient width, the weights' changes during training are small enough to use the Taylor series expansion for the layer outputs. Therefore, inside some small neighborhood of the initial point, $\theta^0$ in the weight space model can be linearized in terms of the weights $\theta$.

\section{Understanding Pre-Trained Weights' Parameter Space}

Having a model pre-trained on some general task (e.g., Language Modeling) $\theta^0 \in \mathbb{R}^{|\theta|}$, it is conventional to initialize a new model with $\theta^0$ when solving a downstream task\footnote{We will refer to superscription $0$, $-$ and $+$ as a sentiment characteristic, similar to how superscription in particle physics refers to particle charge. $0$ refers to a pre-trained model with a neutral sentiment, as we believe that pre-trained models tend to generate texts with a neutral sentiment.}. For example, GPT-2 could be used as $\theta^0$ when training an LM on some specific domain of data (e.g., movie reviews). Doing so makes it possible to obtain faster convergence of training procedures and better results than training from scratch since $\theta^0$ is usually trained with a larger dataset than those available for downstream tasks.

While many works explore the parameter space of models trained from scratch, we are most interested in such a space for fine-tuned models. More specifically, when a model is trained from scratch with different starting points, there is evidence that different $\theta^0$ could be obtained. Furthermore, if different random seeds are used to form mini-batches from the training dataset, additional differences could occur in the resulting parameters of trained models.

It is important to note that, if we train a model from a pre-trained state, we eliminate the randomness caused by different starting points of optimization. From such perspective, we should expect the parameter space of fine-tuned models to be simpler than that of models trained from scratch. To explore the limits of this simplicity, we experimented with linear interpolation between weights of fine-tuned models described in the following sections.

\subsection{Linear Interpolation} \label{linear_interpolation_section}

Consider two models with parameters $\theta^+ \in \mathbb{R}^{|\theta|}$ and $\theta^- \in \mathbb{R}^{|\theta|}$. Both $\theta^+$ and $\theta^-$ are obtained after fine-tuning a pre-trained model $\theta^0$. For convenience, let us consider that $\theta^0$ is Language Model trained on general domain data (e.g., GPT-2), $\theta_+$ and $\theta_-$ are Language Models fine-tuned on positive and negative sentiment data (e.g., SST dataset \citep{sst}). We could linearly interpolate between them as
\begin{equation}
\label{interpolation-eq-1-simple}
    g_1(\alpha) = \alpha \theta^+ + (1 - \alpha) \theta^-, g_1: R \to R^{|\theta|}.
\end{equation}

We can also rewrite $g_1(\alpha)$ differently:

\begin{equation}
\label{interpolation-eq-1}
    g_1(\alpha) = \frac{1}{2}(\theta^+ + \theta^-) + \frac{1}{2}(2\alpha - 1)(\theta^+ - \theta^-),
\end{equation}

which could be seen as moving from starting point $\frac{1}{2}(\theta^+ + \theta^-)$ in a direction $(\theta^+ - \theta^-)$. Expressing interpolation with a starting point such as that in Equation \ref{interpolation-eq-1} could be considered too verbose. However, it allows us to derive the second possible formulation of interpolation, for which we replace the starting moving point with $\theta_0$. To simplify the process even further, we use $\alpha^\prime = 2\alpha - 1$ as an interpolation weight\footnote{Note that we have a different scale for $\alpha^\prime$, since $\alpha = 0$ implies $\alpha^\prime = -\frac{1}{2}$, and $\alpha = 1$ implies $\alpha^\prime = \frac{1}{2}$.}.

\begin{equation}
\label{interpolation-eq-2}
    g_2(\alpha^\prime) = \theta^0 + \alpha^\prime(\theta^+ - \theta^-).
\end{equation}

Going even further, we can decompose the $(\theta^+ - \theta^-)$ direction into $(\theta^+ - \theta^0)$ and $(\theta^- - \theta^0)$, obtaining new parametrization
\begin{equation} \label{interpolation-eq3}
g_3(\alpha, \beta) = \theta^0 + \alpha (\theta^+ - \theta^0) + \beta (\theta^- - \theta^0).
\end{equation}
Note that $\alpha + \beta = 1$ reparametrizes $g_1$ and $\alpha + \beta = 0$ reparametrizes $g_2$.

We discuss the limits of these reparametrizations in the Experiments section.


\subsection{Ensembling} \label{sec:ensembling}

Another way to utilize several models at once is to combine them into an ensemble. While linear interpolation is performed in the weight space, ensembling can be seen as interpolating in the model output space. At every step, language models yield logits $z$ for every token in vocabulary. As proposed in DExperts \cite{liu-etal-2021-dexperts}, we could use these logits to obtain the final tokens' probability:

\begin{equation*}
    \begin{aligned}
        &P(x_t|x_{<t}) = \operatorname{softmax(z)}, \\
        &z = z^0(x_{<t}) + \alpha \cdot (z^+(x_{<t}) - z^-(x_{<t}))
    \end{aligned}
    \label{eq:logits_int}
\end{equation*}

However, this method requires significant computational time overhead compared to interpolation in the parameter space since it requires evaluating several models to get predictions.

\begin{figure*}[h!]
  \centering

  \medskip
    \begin{subfigure}[t]{0.32\linewidth}
    \centering\includegraphics[width=\linewidth]{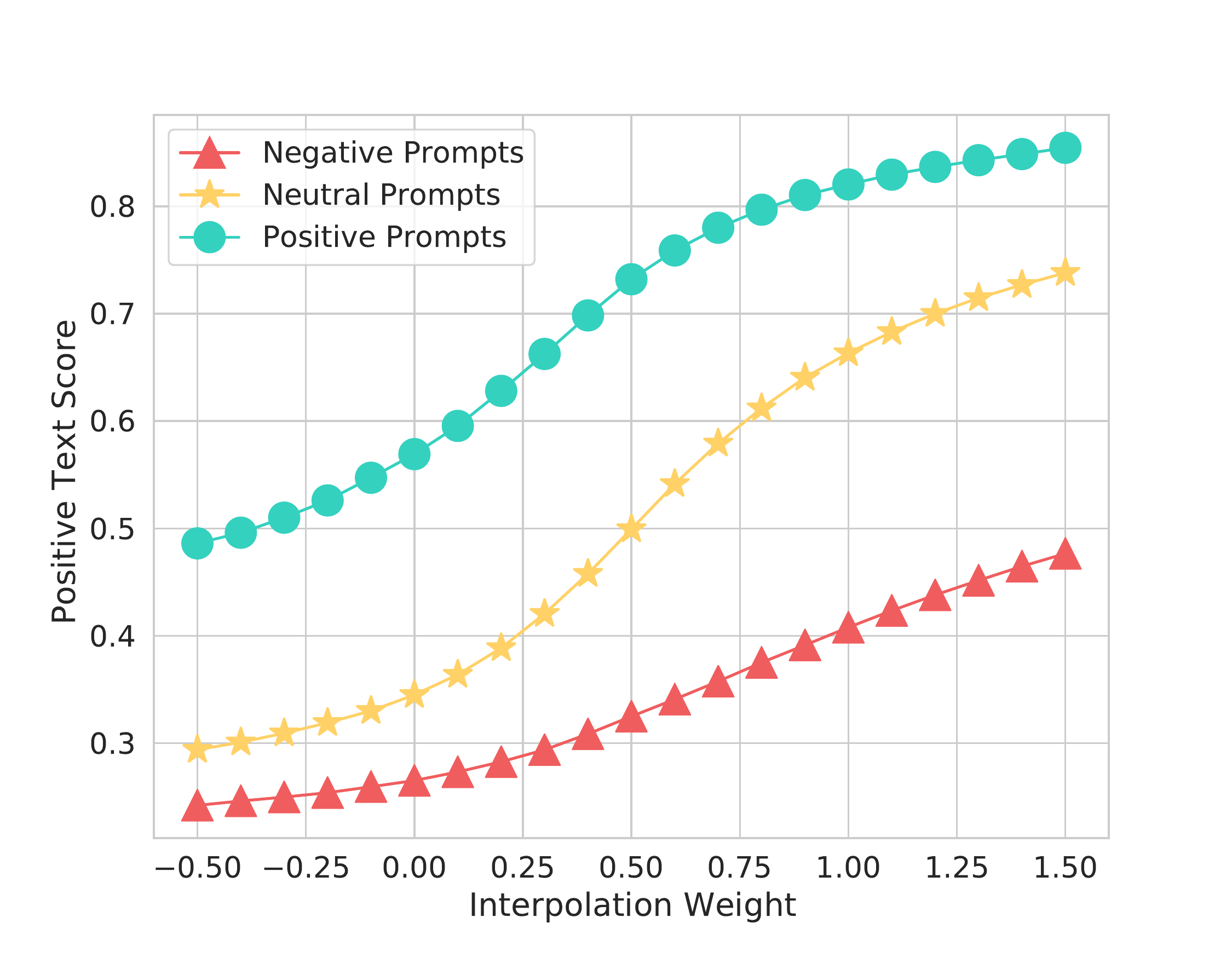}
    \caption{}
  \end{subfigure}
  \begin{subfigure}[t]{0.32\linewidth}
    \centering\includegraphics[width=\linewidth]{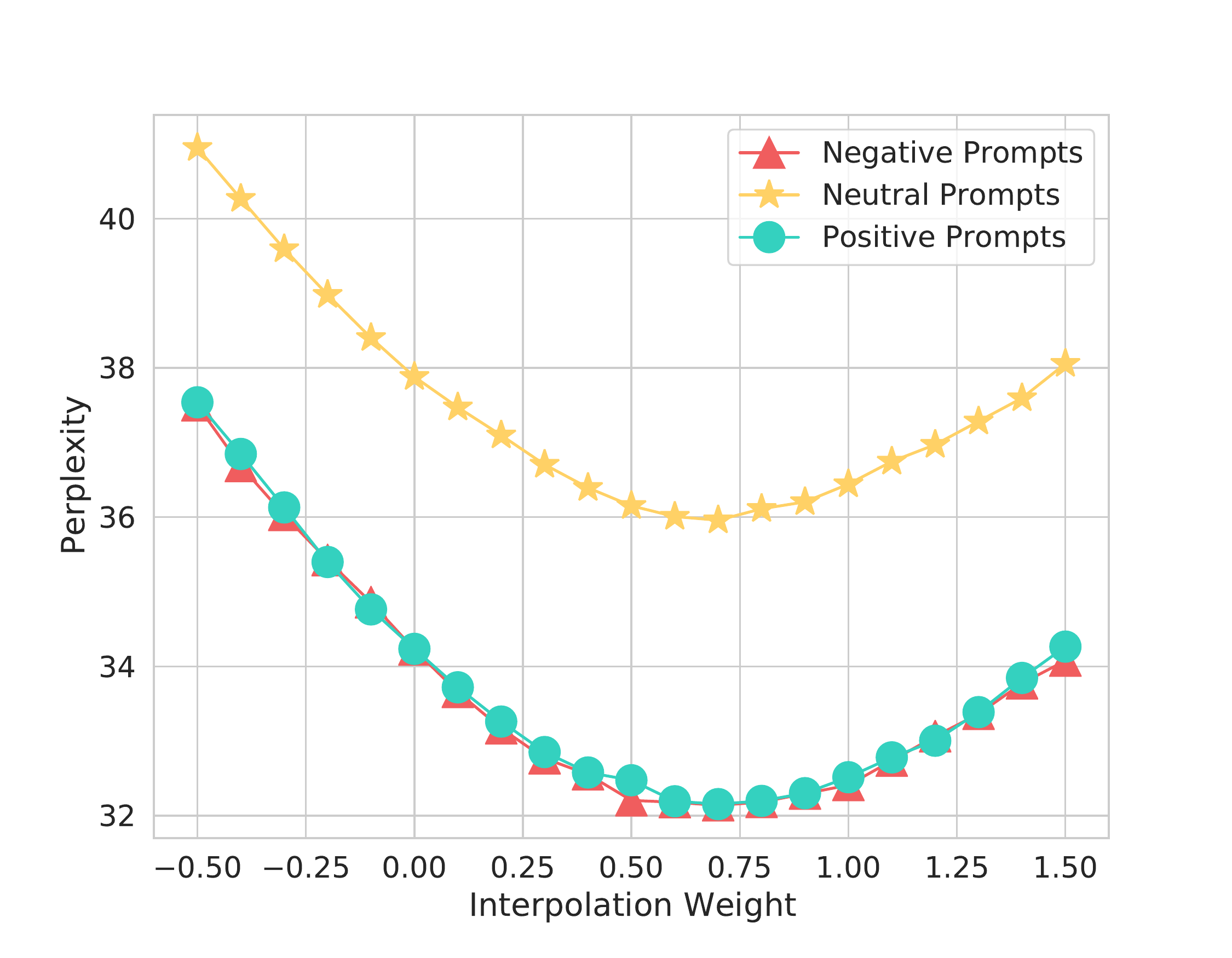}
    \caption{}
  \end{subfigure}
  \begin{subfigure}[t]{0.32\linewidth}
    \centering\includegraphics[width=\linewidth]{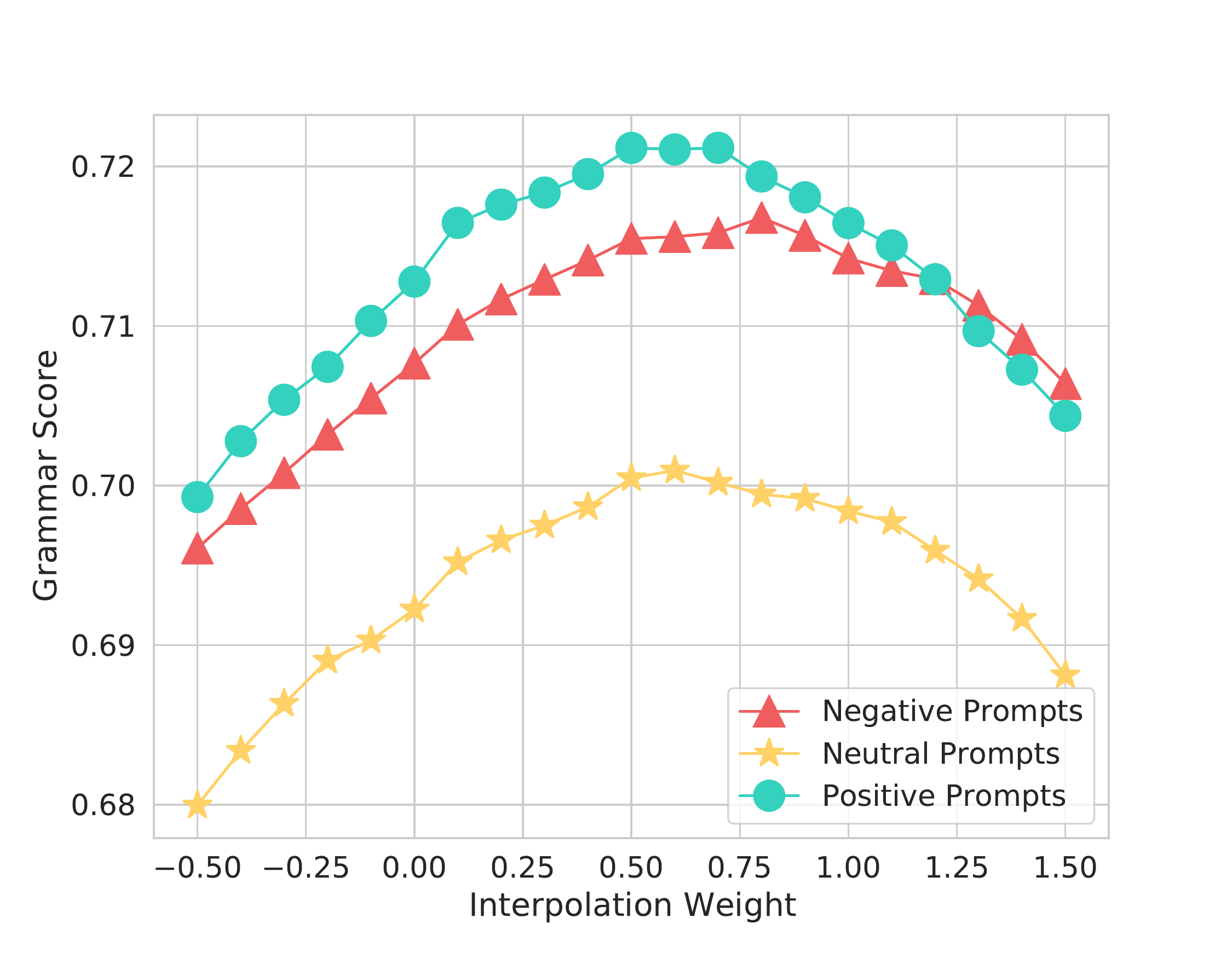}
    \caption{}
  \end{subfigure}
  \caption{Interpolation between two models fine-tuned on positive and negative sentiment with $g_1(\alpha)$. We report the Mean probability of the positive sentiment (a), the Perplexity of the generated text (b), and the Probability of the grammatically correct text (c) for obtained interpolated models. See Section \ref{ctg_section} for more details.}
  \label{fig:res_contr_text_gen}
 \end{figure*}

\section{Experiments} \label{sec:experiments}

\begin{figure}[h!]
    \centering
    \includegraphics[width=\linewidth]{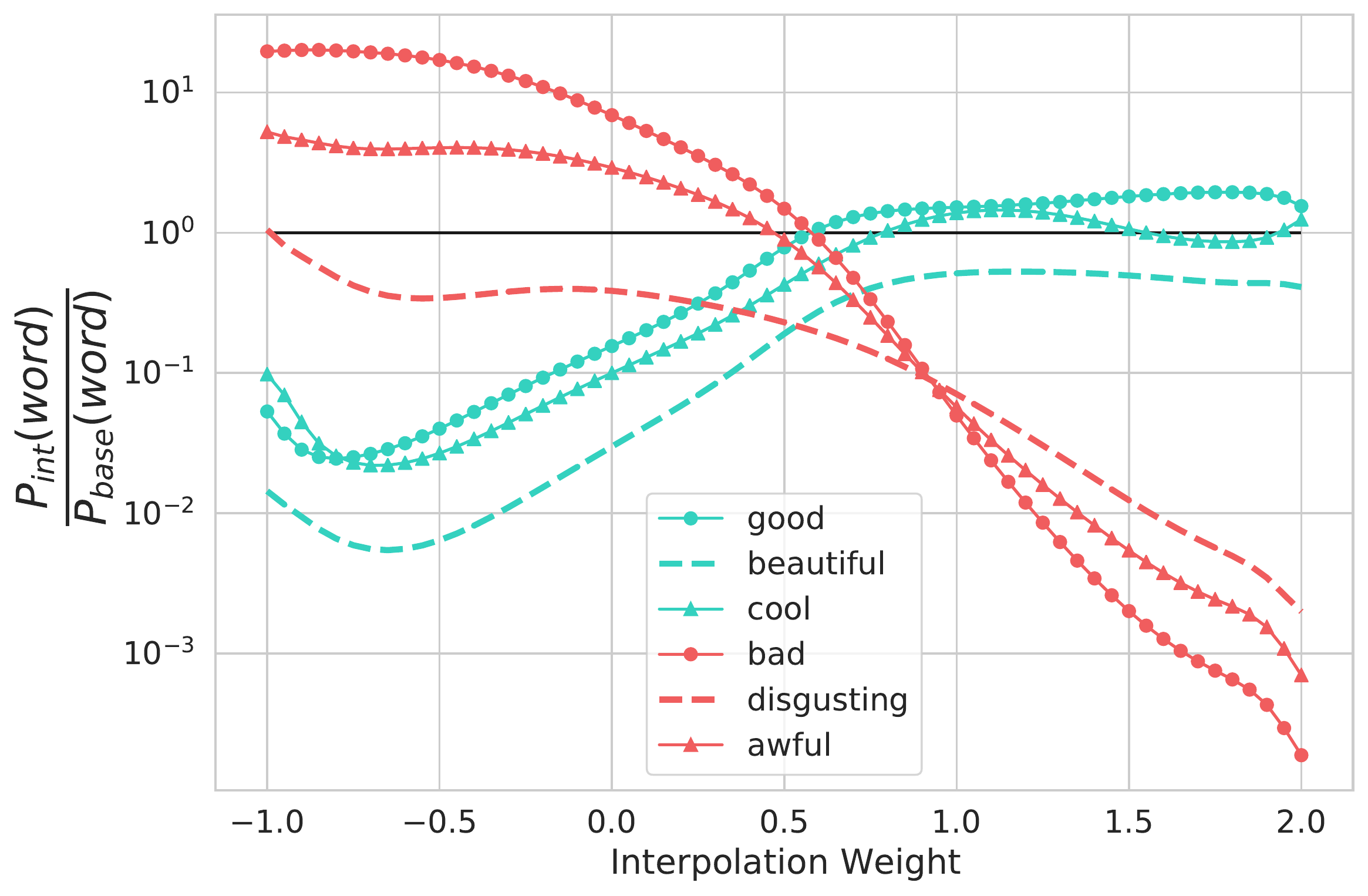}
    \caption{Words' probability during interpolation with the text prompt "The movie was". Words leading to positive sentiment are plotted in green, and negative in red. While interpolating between $\theta^-$ and $\theta^+$, probabilities of negative words decrease, and positive ones increase. See \Cref{ctg_section} for more details.}
    \label{fig:words_prob}
\end{figure}

\subsection{Controllable Text Generation} \label{ctg_section}

Controllable text generation can be seen as the simplest way to explore the parameter space of fine-tuned models. The performance of obtained $\theta$ can be quickly evaluated with automatic metrics such as desired attribute probability of generated texts, and text quality can be evaluated by perplexity and grammar correctness. 

Following the DExperts \cite{liu-etal-2021-dexperts} setup, we took the SST dataset containing texts with labels representing the sentiment of sequences. We constructed a positive sentiment dataset containing texts with labels such as "positive" or "very positive". In addition, we also created a negative dataset with "negative" and "very negative" texts. We then fine-tuned two GPT-2 Large models on the causal language modeling task on these datasets to obtain $\theta^+$ and $\theta^-$, respectively. 

We then evaluated the models obtained by $g_1$ and $g_2$ (See Equations \ref{interpolation-eq-1-simple},  \ref{interpolation-eq-2}) to understand the limits of linear interpolation for fine-tuned models. To do so, we used the same prompts as DExperts \cite{liu-etal-2021-dexperts} for text generation. For every prompt, we generated 25 continuations with their length less or equal to 30 tokens. 

We used three metrics to evaluate the generated texts' sentiment and quality. Positive text scores are evaluated using an external classifier and show the mean probability of positive sentiment in the generated text. Grammar scores are determined by a classifier trained on the CoLA \cite{glue} dataset. To evaluate the texts' quality, we calculate perplexity using GPT-2 XL. 

See Section \ref{sec:det_controllable_text_generation} of the  Appendix for more details on training and evaluation used in these experiments.

See Figure \ref{fig:res_contr_text_gen} for the results. We found that perplexity with $g_1(\alpha)$ remains stable in $\alpha \in [0;1]$, in which we have a zero perplexity barrier. A wider interval of $\alpha$ also shows promising results, where the positive sentiment probability increases with $\alpha > 1$. Meanwhile, perplexity and grammar remain stable. Based on this, we can assume that models obtained by simple linear interpolation can still be considered language models. Moreover, the original's features, such as positive sentiment probability, could be enhanced by $\alpha > 1$ (and vice versa). In Section \ref{sec:analyzing_results}, we hypothesise why linear interpolation in the weight space works even in cases of complex non-linear models.

\begin{figure*}[h!]
  \centering

  \medskip
    \begin{subfigure}[t]{0.3\linewidth}
    \centering\includegraphics[width=\linewidth]{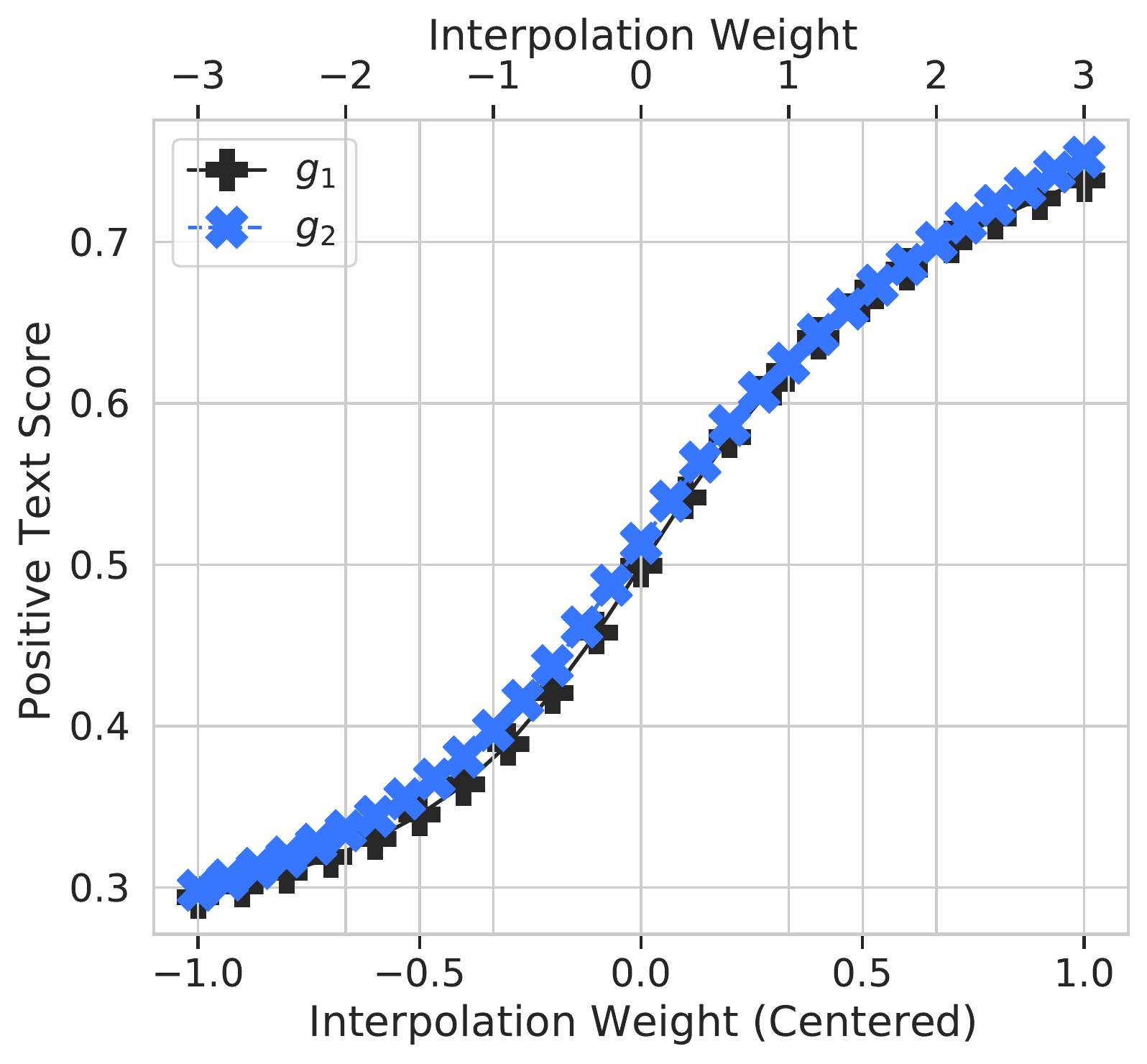}
    \caption{}
    \label{fig:1_vs_2_a}
  \end{subfigure}
  \begin{subfigure}[t]{0.3\linewidth}
    \centering\includegraphics[width=\linewidth]{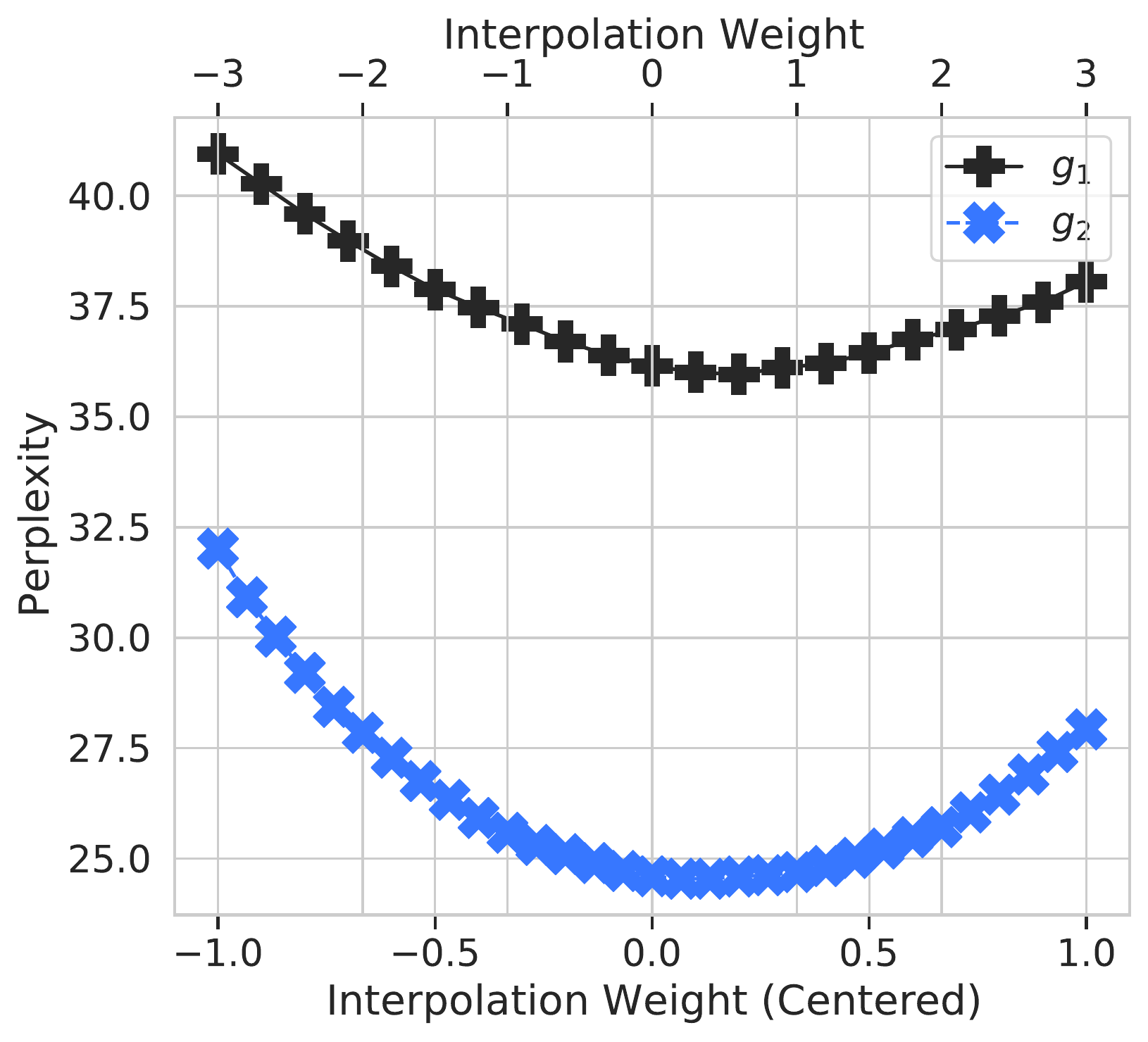}
    \caption{}
  \end{subfigure}
  \begin{subfigure}[t]{0.3\linewidth}
    \centering\includegraphics[width=\linewidth]{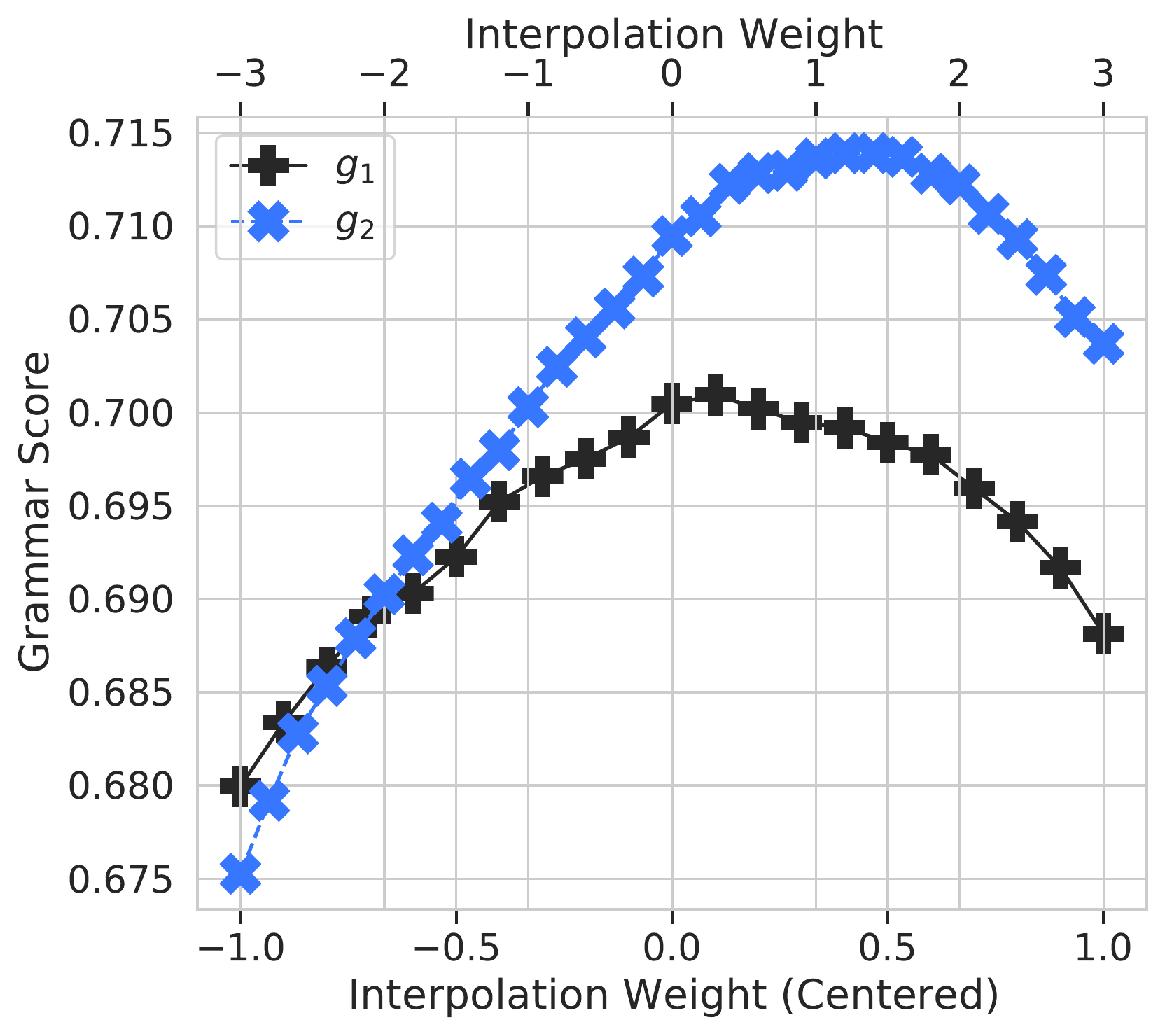}
    \caption{}
  \end{subfigure}
  \caption{Comparison of $g_1(\alpha)$ and $g_2(\alpha)$ interpolations with neutral prompts. We evaluated the positive text score (a), perplexity (b), and Grammar Correctness (c) for interpolated models. Note that these interpolation methods differ in the scale of $\alpha$ (See Section \ref{linear_interpolation_section} for details). Therefore, we used different scales to report these results. $\alpha$ values for $g_1(\alpha)$ are shown below the plot, while those for $g_2(\alpha)$ are above. While the positiveness of both approaches is comparable, $g_2(\alpha)$ obtained better perplexity and grammar correctness through utilizing $\theta^0$ parameters. See Section \ref{sec:1_vs_2} for more details.}
  \label{fig:1_vs_2}
 \end{figure*}
 
 \begin{figure}[h!]
  \centering

  \begin{subfigure}[t]{0.85\linewidth}
    \centering\includegraphics[width=\linewidth]{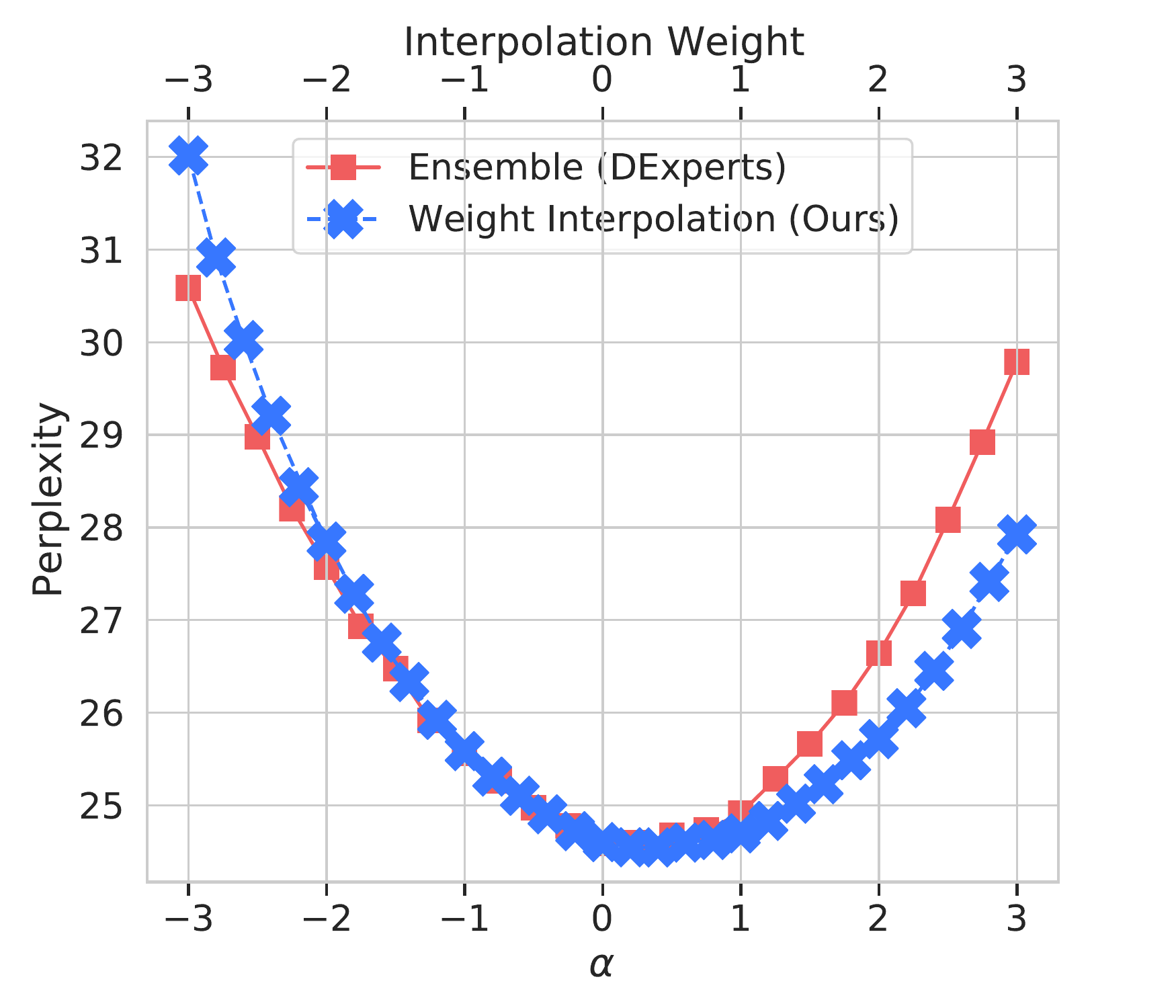}
    \caption{}
  \end{subfigure}
      \begin{subfigure}[t]{0.85\linewidth}
    \centering\includegraphics[width=\linewidth]{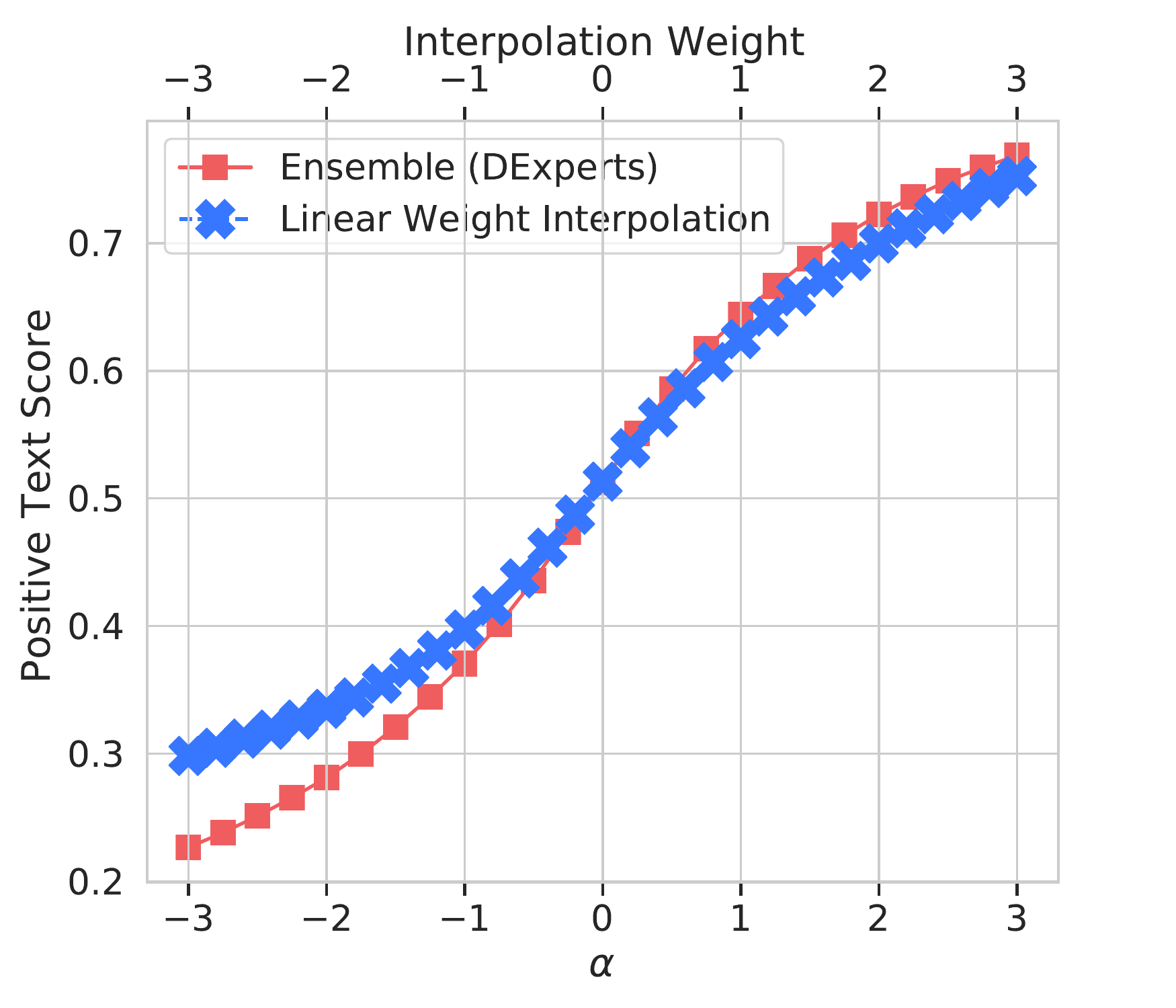}
    \caption{}
  \end{subfigure}
  \caption{Perplexity (a) and Positive Text Score (b) with respect to interpolation weight in the logits space (DExperts) and the weight space. Note that the text's positiveness increases smoothly in booth cases, while perplexity remains low. See Section \ref{sec:experiments} for more details.}
  \label{fig:int_vs_ens}
 \end{figure}

We also measured the next token probability on our "The movie was" prompt using parameters obtained from $g_1$ interpolation. See Figure \ref{fig:words_prob} for the results. The probabilities of the words leading to positive sentiment are monotonically increasing, while the probabilities of the negative sentiment words are monotonically decreasing with $\alpha$.


\subsection{Which Parametrization Is Best?}\label{sec:1_vs_2}

In Section \ref{linear_interpolation_section}, we discussed different ways to parametrize interpolation and move the model's weights in the desired direction. However, note that it is not fully clear what the differences are between them. To compare the proposed interpolation schemes, we conducted experiments with the pre-trained model GPT-2 Large as $\theta^0$ and Fine-Tuned models from the previous section as $\theta^+$ and $\theta^-$. Note that parametrization $g_2$ takes three models as input.  

\begin{figure}[h!]
  \centering
    \begin{subfigure}[t]{0.99\linewidth}
    \centering\includegraphics[width=\linewidth]{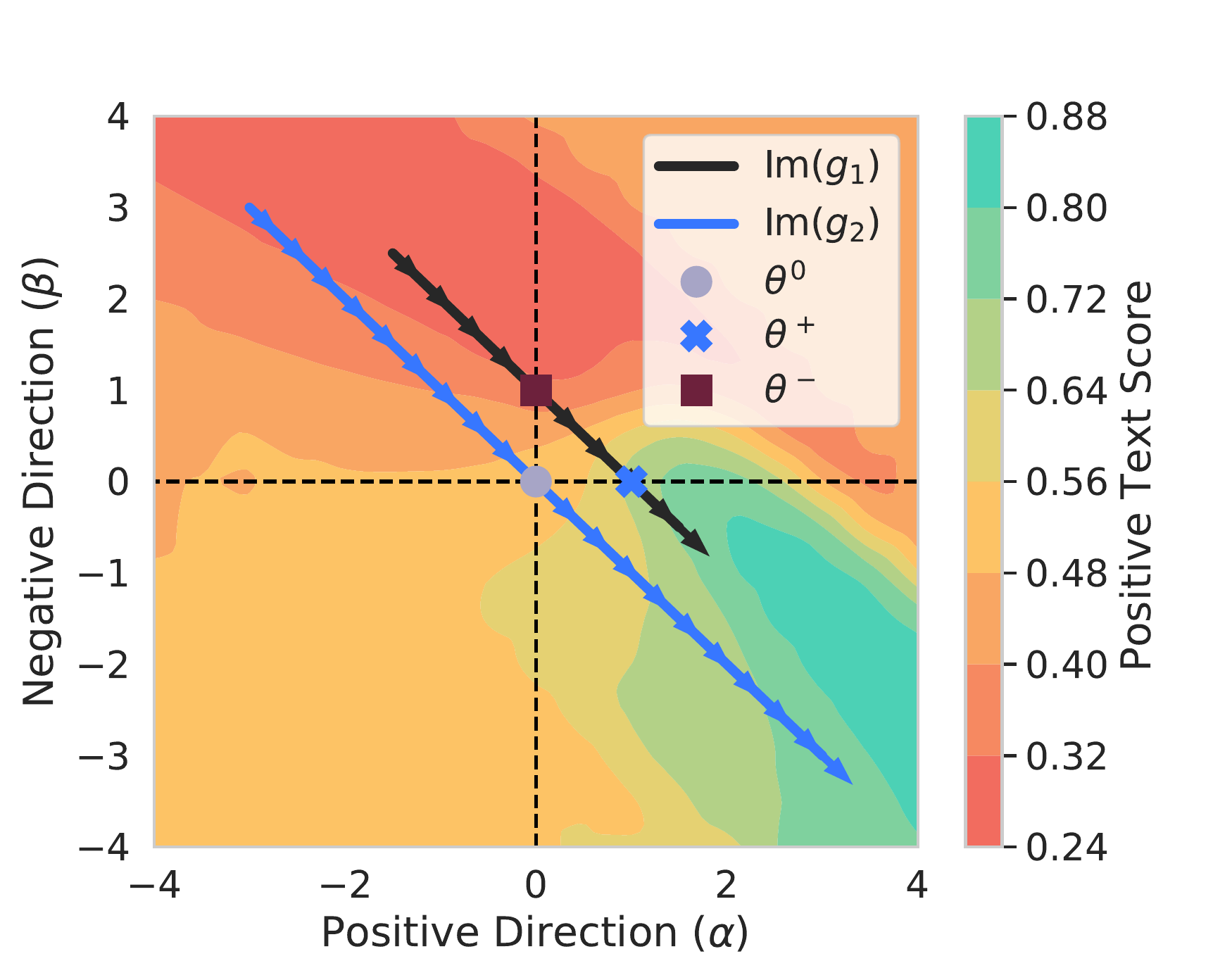}
    \caption{}
  \end{subfigure} \qquad
  \begin{subfigure}[t]{0.99\linewidth}
    \centering\includegraphics[width=\linewidth]{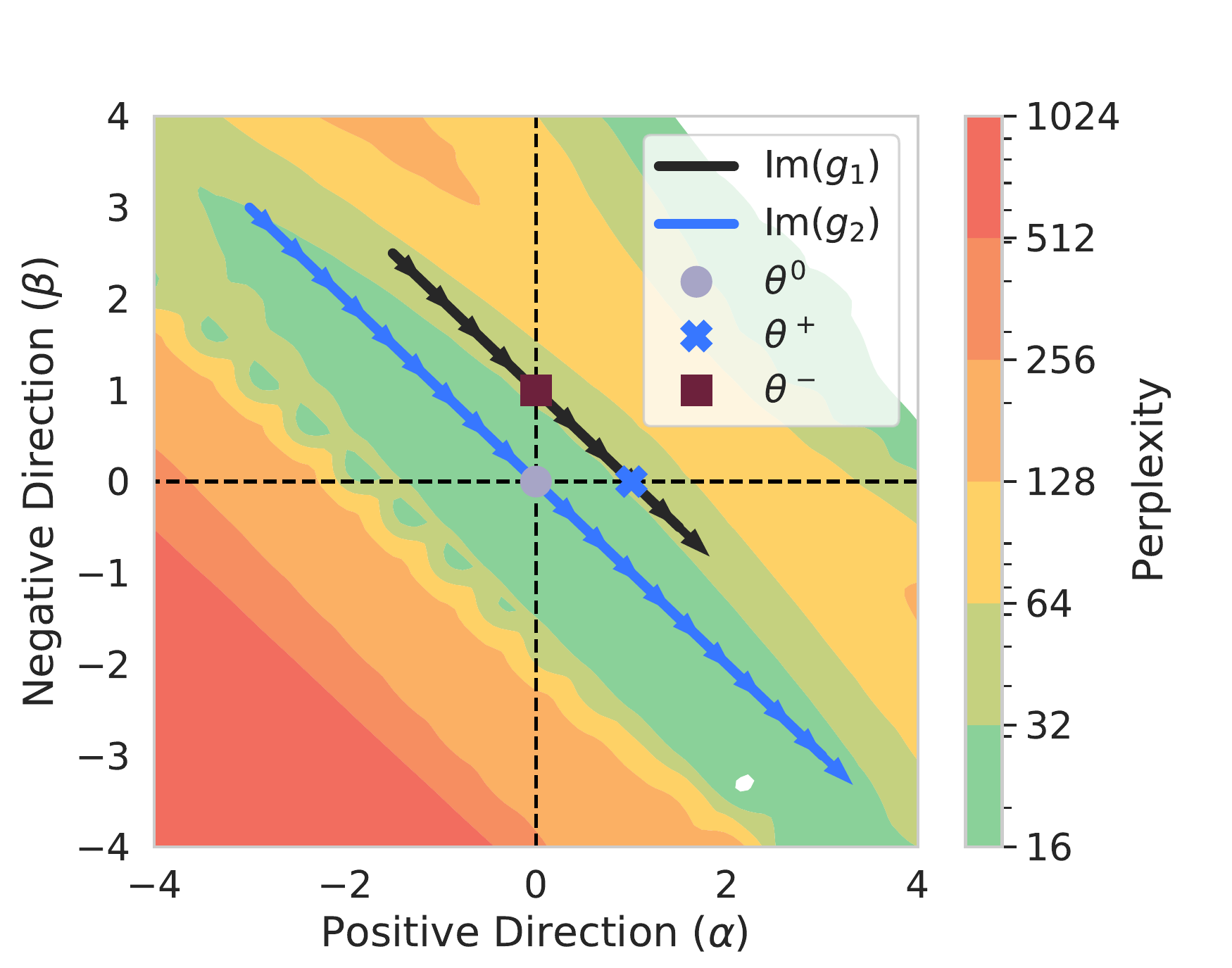}
    \caption{}
  \end{subfigure}
  \caption{Postive text score (a) and perplexity (b) for $g_3(\alpha, \beta)$ interpolation. Positive text score has a clear growing direction from upper left to lower right. We also found plateau of perplexity near $\alpha = -\beta$ line. See Section \ref{sec:interpolation_space} for more details.}
  \label{fig:g_3_perp_attr}
 \end{figure}

In this subsection, all experiments were conducted with neutral prompts only. See Figure \ref{fig:1_vs_2} for the results. We observed that $g_1(\alpha)$ obtained a positiveness score comparable with $g_2(\alpha)$, while the latter showed better perplexity and grammar correctness. However, we would like to note that, in this case, perplexity should not be considered fully representative of the generated texts' quality. Since $g_2(\alpha)$ utilizes $\theta^0$, its samplings are more likely to produce texts which would be treated as more probable by GPT-2, while $g_1(\alpha)$ has a stronger shift towards movie reviews. 

\subsection{Interpolation Space}\label{sec:interpolation_space}

\begin{figure}[h!]
  \centering
    \begin{subfigure}[t]{0.99\linewidth}
    \centering\includegraphics[width=\linewidth]{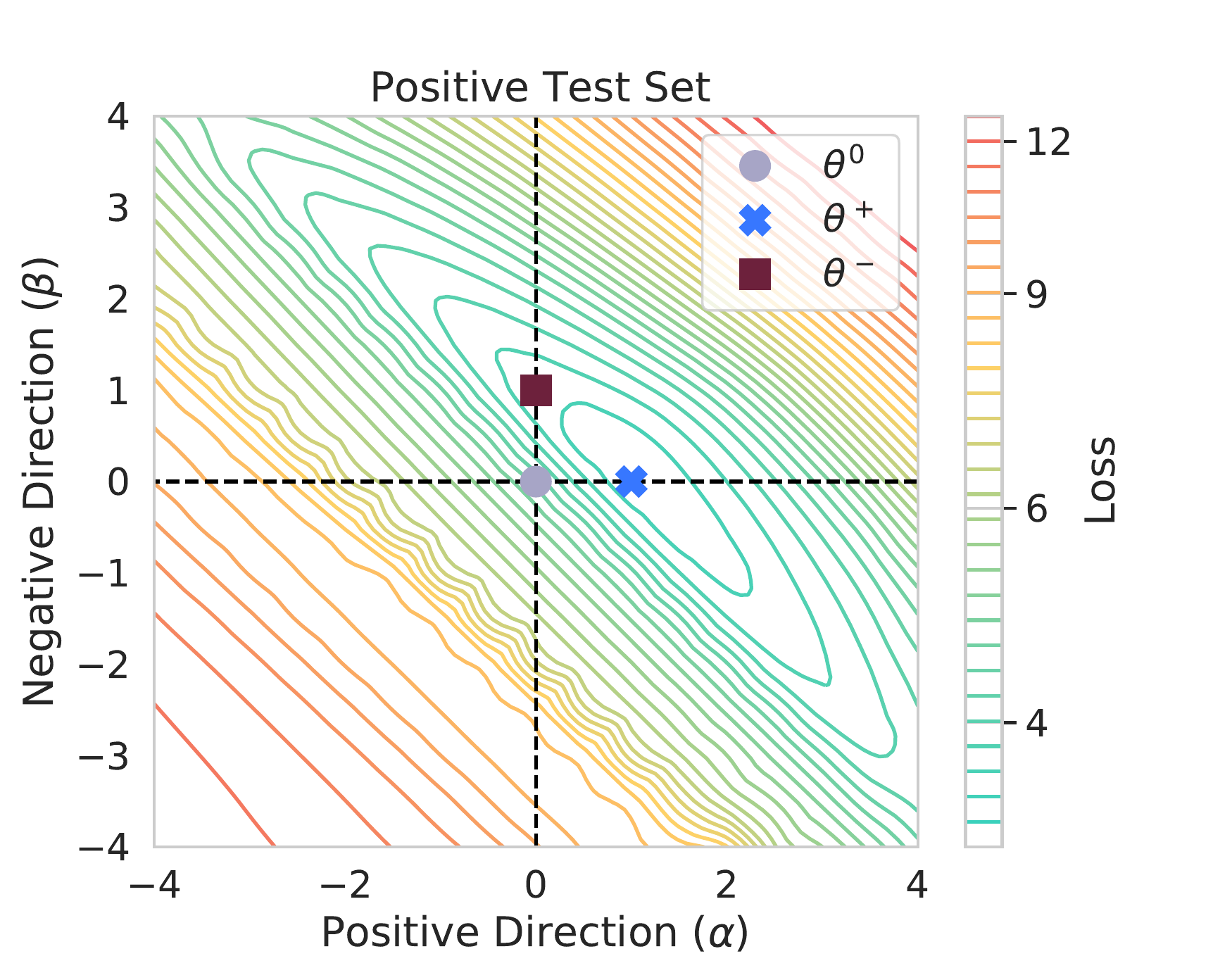}
    \caption{}
  \end{subfigure} \qquad
  \begin{subfigure}[t]{0.99\linewidth}
    \centering\includegraphics[width=\linewidth]{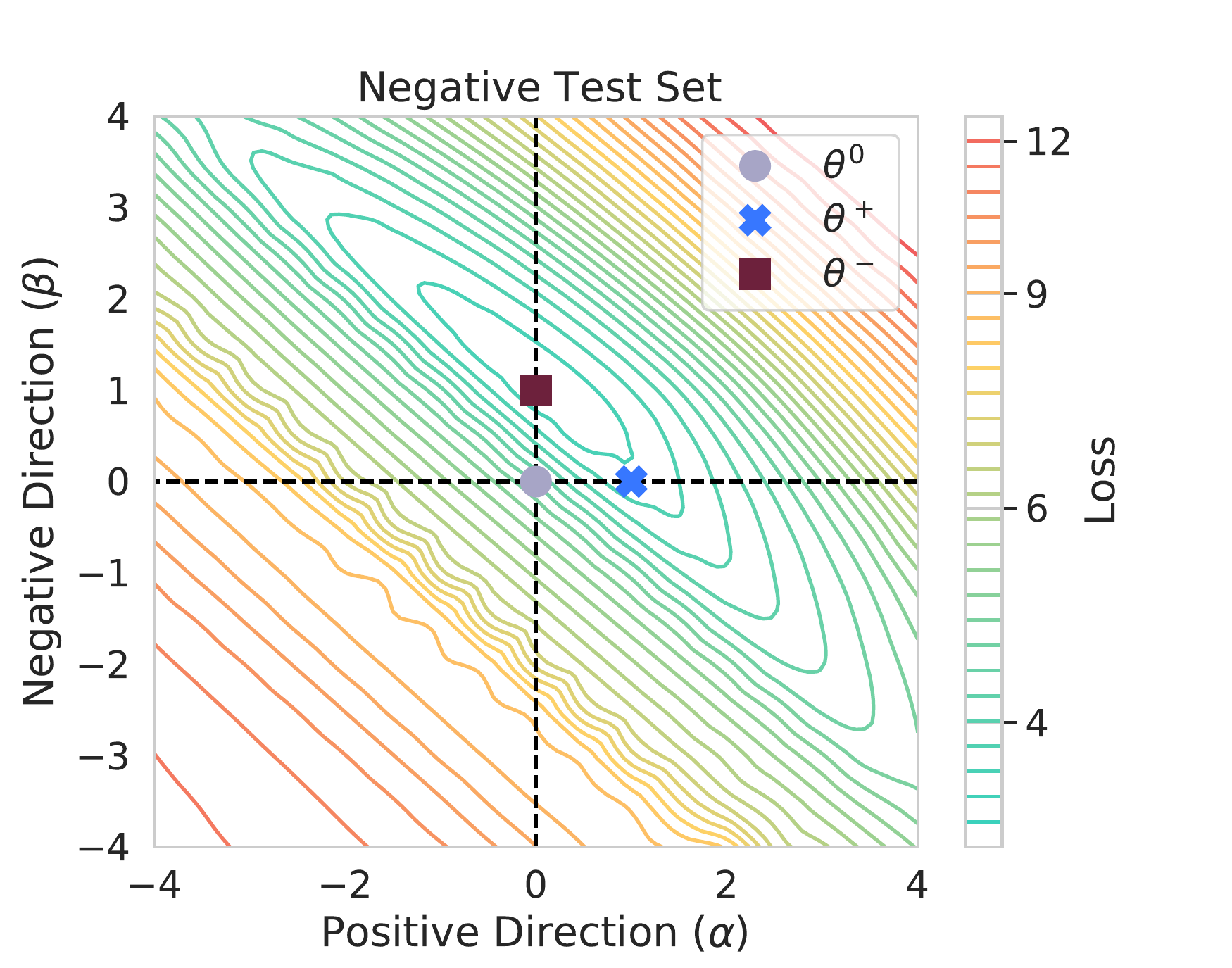}
    \caption{}
  \end{subfigure}
  \caption{Positive (a) and negative (b) test set losses for the models obtained by $\theta = g_3(\alpha, \beta)$ interpolation. See Section \ref{sec:interpolation_space} for more details.}
  \label{fig:nll_test}
 \end{figure}

To further analyze the interpolation points, we conduct experiments with interpolation $\theta = g_3(\alpha, \beta)$. Note that $g_3: \mathbb{R}^2 \to \mathbb{R}^{|\theta|}$ maps point $(0, 1)$ to $g(0, 1) = \theta^0 - \theta^0 + \theta^+ - 0 \cdot(\theta^- - \theta^0) = \theta^+$, analogously $(0, 0) \to \theta^0$ and $(1, 0) \to \theta^-$. Then we can choose any point in $\mathbb{R}^2 = \operatorname{dom}g$ and obtain a model $\theta = g(\alpha, \beta)$. We use a 2d uniform grid with values from -4 to +4 and 20 points in every dimension ($G = \{i/2.5\}_{i=-10}^{10} \times \{j/2.5\}_{j=-10}^{10}$) to obtain 400 models, and measure the properties of these models. As a result, we get 400 points of perplexity and a positive sentiment score shown in Figure \ref{fig:g_3_perp_attr}. In addition, we also count the models' mean negative log-likelihood loss value on a test set of SST positive and negative subsets. See Figure \ref{fig:nll_test} for the results. 

We found a plateau of perplexity near the $\alpha = -\beta$ line. Furthermore, we also found that the loss values of models near $\alpha = -\beta + 1$ (equals to $\theta = g_2(\cdot)$ parametrization) are significantly lower. These results can be explained as follows. The first parametrization $\theta = g_1(\cdot)$ does not utilize a pre-trained $\theta^0$ model. Therefore, the obtained models remain within the SST dataset domain (movie reviews). We can assume that it is because of lower loss on both test subsets. The perplexity of the specific texts is higher due to them becoming biased toward the movie review style. The model we used to measure perplexity (pre-trained GPT-2 XL) is not a domain-specific model and therefore measures information contained in the generated text. As the domain of the generated texts shifted, we observed consistently higher perplexity compared to $\theta = g_2(\cdot)$. However, this perplexity remained stable and in a meaningful value below 50. On the other hand, the parametrization $\theta = g_2(\cdot)$ did not shift toward the movie reviews domain because of the constant persistence of the $\theta^0$ term. Lower perplexity, in this case, did not indicate better quality of the generated texts.

\subsection{Interpolation vs. Ensembling} \label{sec:int_vs_ens}

In this section, we compare two methods of utilizing several models for controllable text generation tasks.

 As discussed in \ref{sec:ensembling}, DExperts could be seen as a linear interpolation in the model outputs space.
 
 We generated texts with several values of the $\alpha$ parameter. Then, we used $\theta^{+}$ model as an expert and $\theta^-$ model as an anti-expert. We compared this setup with $\theta = g_2(\alpha)$ parametrization. The results are presented in Figure \ref{fig:int_vs_ens}.
 
 We found that the curves are almost identical for $\alpha \in [0;1]$. Similar results could be obtained if all model backbones were linear. Surprisingly for us, we also discovered that linear interpolation in weight space is highly competitive to ensembling and does not damage the internal knowledge of the model.

\section{Results Analysis} \label{sec:analyzing_results}

  \begin{figure*}[t!]
  \centering

  \medskip
    \begin{subfigure}[t]{0.368\linewidth}
    \centering\includegraphics[width=\linewidth]{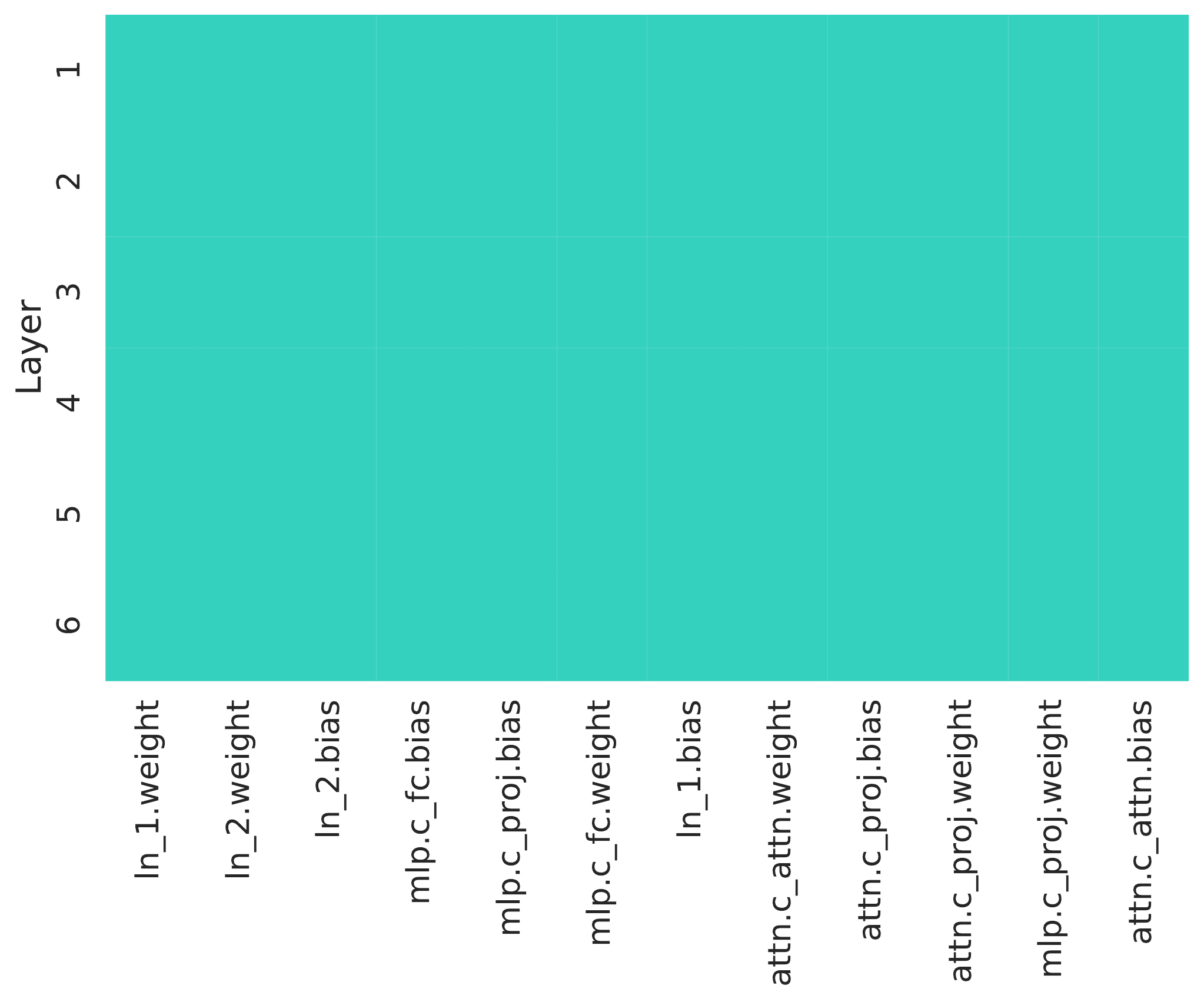}
    \caption{}
  \end{subfigure}
  \begin{subfigure}[t]{0.425\linewidth}
    \centering\includegraphics[width=\linewidth]{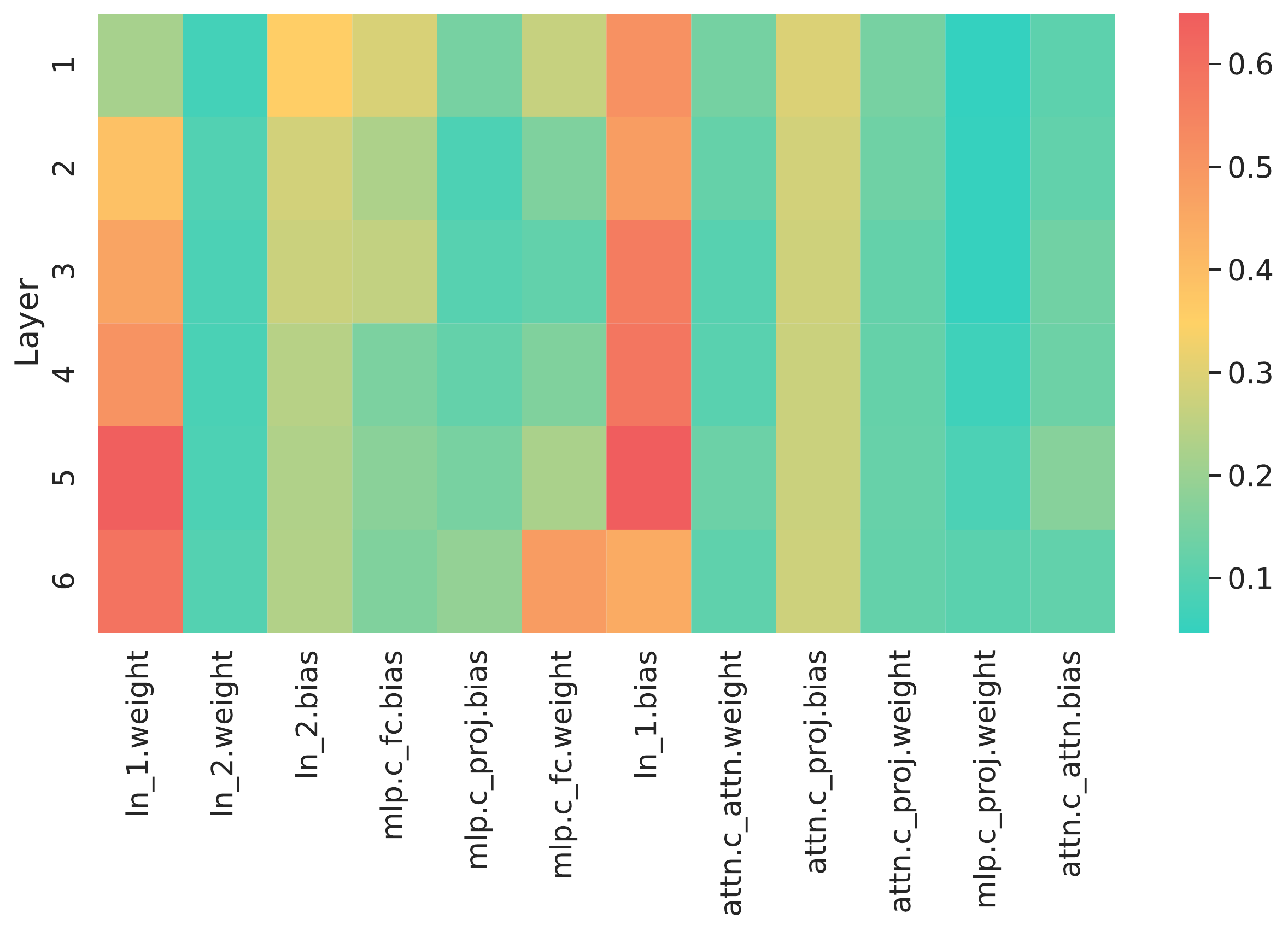}
    \caption{}
  \end{subfigure}
  \caption{The norm of the weight difference for $\theta^+$ and $\theta^0$ (a), as well as DistilGPT-2 trained on C4 \cite{t5} from scratch and pre-trained DistilGPT-2 (b). Each row represents one layer, and the parameter names can be found on the x-axis ticks. See Section \ref{sec:c4_lm} for more details.}
  \label{fig:delta_w}
 \end{figure*}

To explain the results of the above observations, we will now try to establish some intuition on why linear interpolation works so well for pre-trained language models.

\subsection{Lazy Training}

Lazy training, introduced by \citealt{lazy_training}, has a solid connection to linear interpolation in the weight space.

Let us say that we have some function $f(\theta) = R(h(\theta)): \mathbb{R}^{|\theta|} \to \mathbb{R}_+$, where $h(\theta): \mathbb{R}^{|\theta|} \to \mathcal{F}$ is our model.
Now, let us define a linearized model $\overline{h}(\theta) = h(\theta^0) - D h(\theta^0) (\theta^0 - \theta)$. $\overline{h}(\theta)$ could be seen as a Taylor's series expansion to the first order of $h(\theta)$. If we can accurately approximate $h(\theta)$ with $\overline{h}(\theta)$, then $\overline{f}(\theta) = R\left(\overline{h}(\theta)\right)$ becomes a good approximation of $f$ if we are using gradient descent for optimizing $R$.

\subsection{Pre-Trained Models Fine-Tuning and Lazy Traing}

\citealt{lazy_training} considers function $f$ to be a loss function optimized by gradient descent. In our work, we will talk about a proxy function that can be seen as a differentiable interpolation of the positive text score or other desired attributes for controllable text generation. For example, we can start our optimization process at $\theta^0$ and train models $\theta_+$ and $\theta_-$ using stochastic gradient descent on the NLL target function. We believe that after minimization, some function $f$ (in our case, the probability of positive sentiment in the generated text) will have a lower value on $\theta^-$ and a higher value on $\theta^+$. In other words, during the training procedure, we are trying to find weights $\theta^-$ and $\theta^+$ such that $f(\theta^-) < f(\theta^0) < f(\theta^+)$.

\textbf{Conjecture.} \textit{Point $\hat{\theta}$ with the lowest value of the loss function $L$, such as NLL, does not imply optimal value of the truly desired $f$ function. In other words, we can find a value of $\theta^*$ with $L(\theta^*) > L(\hat{\theta})$, but $f(\theta^*) > f(\hat{\theta})$.}

Function $f$ can be a composition of other functions such as a weighted sum of grammar scores, desired and present attributes, and perplexity.

\textbf{Assumption. }\label{as:as}\textit{If the weights $\theta$ obtained after the fine-tuning procedure are close to pre-trained initialization $\theta^0$, we can linearize the function $f$ as $\overline{f}(\theta) = f(\theta^0) + \nabla f(\theta^0)^T (\theta - \theta_0)$ in some neighbourhood of $\theta^0$.}

If we then parametrize $\theta$ with the general parametrization $\theta = g_3(\alpha, \beta) = \theta^0 + \alpha(\theta^+ - \theta^0) + \beta(\theta^- - \theta^0)$ and pass it to $\overline{f}$, we obtain

\begin{equation}
    \begin{aligned}
    & \overline{f}\circ g(\alpha, \beta) = f(\theta^0) \\
                                       &+ \alpha \cdot \nabla f(\theta^0)^T (\theta^+ - \theta^0) \\
                                       &+ \beta \cdot \nabla f(\theta^0)^T (\theta^- - \theta^0)  \\
                                       &= \alpha \cdot C^+ + \beta \cdot C^-,
    \end{aligned}
\end{equation}
where $C^+$ and $C^-$ are constants.

Note that $C^+ \approx \partial (f \circ g) / \partial \alpha$ and $C^- \approx \partial (f \circ g) / \partial \beta$ in some $\theta^0$ neighbourhood.

The scheme with linear interpolation works even if $C^+$ and $C^-$ are not constants, since $C^+(\alpha)$ > 0 and $C^-(\beta) < 0$ is a sufficient condition.

This model clarifies the similarity between DExperts and linear weight interpolation in \Cref{fig:int_vs_ens}. If Assumption \ref{as:as} holds, then the interpolation between weights will be approximately equal to the interpolation between outputs in a small enough region around $\theta^0$

  \begin{figure*}[t!]
  \centering

  \medskip
    \begin{subfigure}[t]{0.49\linewidth}
    \centering\includegraphics[width=\linewidth]{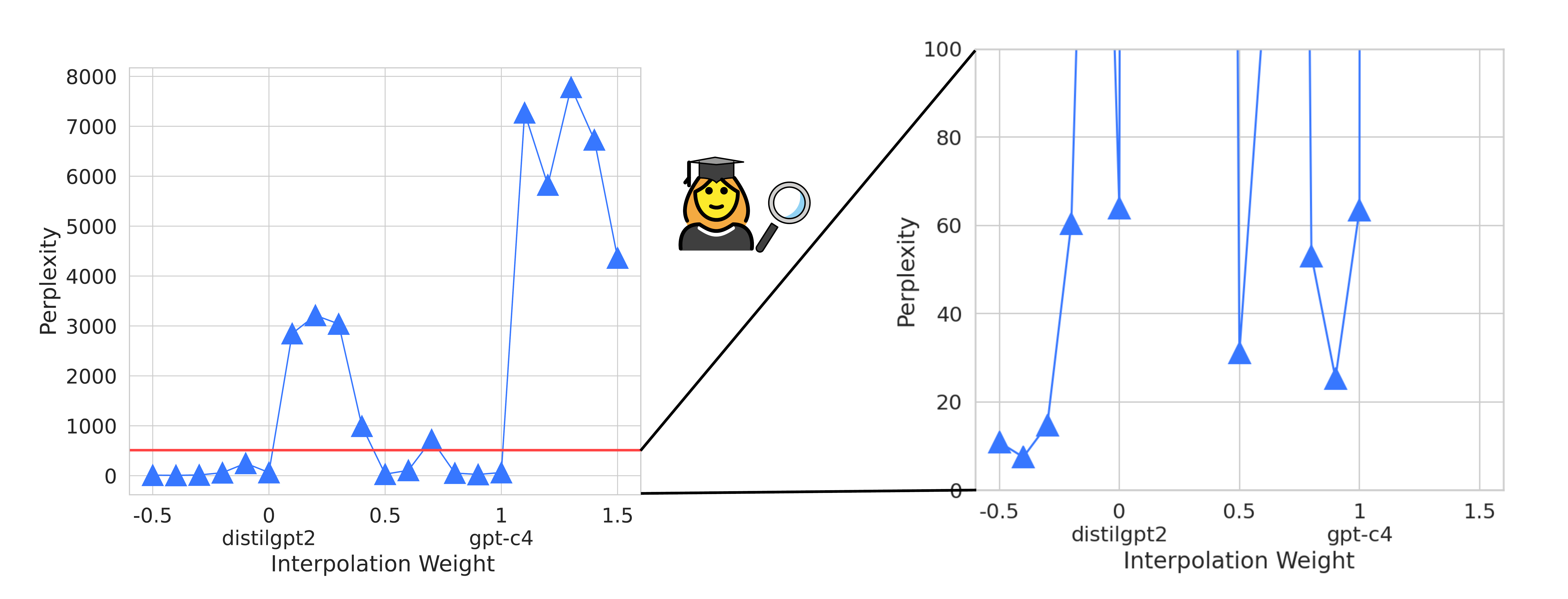}
    \caption{}
    \label{fig:distil_to_c4_perplexity}
  \end{subfigure}
  \begin{subfigure}[t]{0.24\linewidth}
    \centering\includegraphics[width=\linewidth]{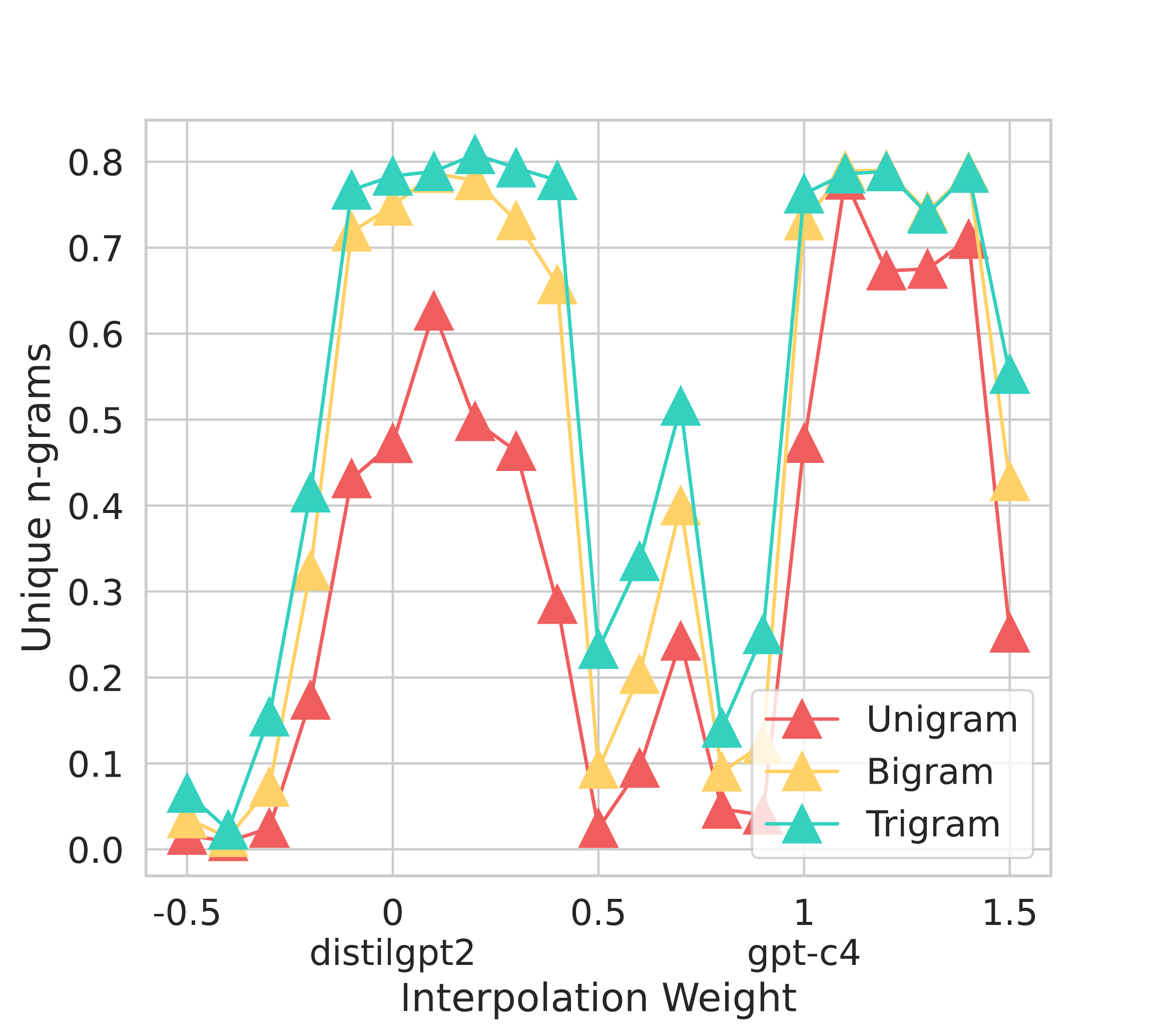}
    \caption{}
    \label{fig:distil_to_c4_n_grams}
  \end{subfigure}
    \begin{subfigure}[t]{0.24\linewidth}
    \centering\includegraphics[width=\linewidth]{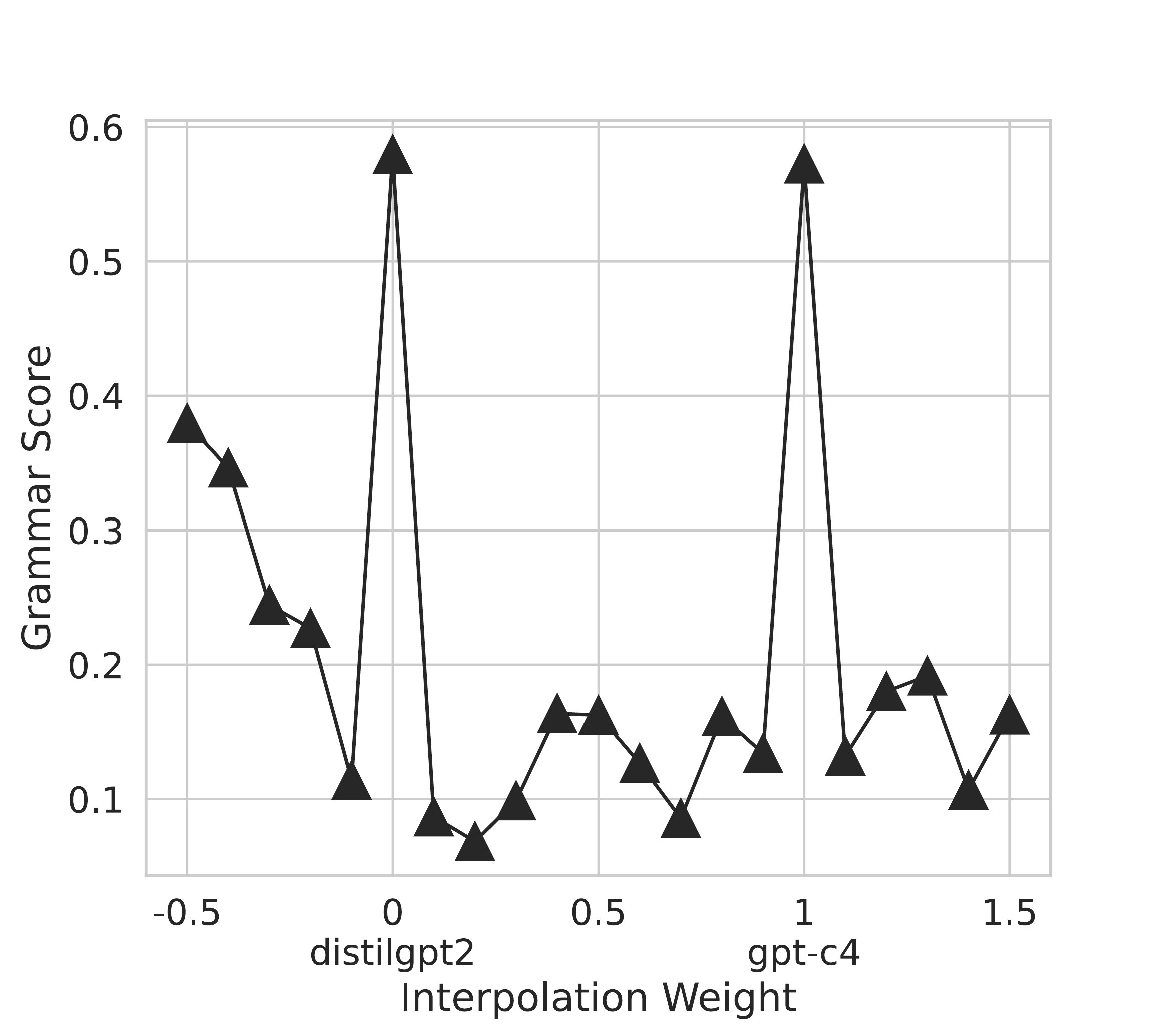}
    \caption{}
    \label{fig:distil_to_c4_grammar}
  \end{subfigure}
  \caption{
Linear interpolation between DistilGPT-2 and GPT-C4 weights. We observed that points between original weights performed poorly with increased perplexity (a), the reduced fraction of unique n-grams (b), and grammar score (c). See Section \ref{sec:c4_lm} for more details.}
  \label{fig:distil_to_c4}
 \end{figure*}

\subsection{Interpolating Between Two Different Decorrelated Language Models} \label{sec:c4_lm}

The small difference between weights is the main factor for the above-mentioned theory. We hypothesize that we obtained such a low difference since we performed fine-tuning of the same pre-trained model $\theta^0$. To support this, we conduct experiments with two different language models.

For the simplicity of our experiment, we chose a small DistilGPT-2 model \cite{sanh2019distilbert} with 6 hidden layers. We trained a GPT model from scratch on C4 \cite{2019t5}, namely \textbf{GPT-C4} with architecture identical to DistilGPT-2. For more details, refer to \Cref{sec:lm_training_details}.

Firstly, we measured the norm of the weight differences between two pairs of models.

\begin{enumerate}[(a)]
    \item Pre-trained original DistilGPT-2 $\longleftrightarrow$ Fine-tuned original DistilGPT-2 on positive sentiment.
    \item Pre-trained original DistilGPT-2 $\longleftrightarrow$ Pre-trained on C4 dataset GPT-C4.
\end{enumerate}

To evaluate the difference between 1-d tensors (biases), we use the scaled $\ell^2$-norm:
$$
\Delta_b = \frac{\lVert b_1 - b_2 \rVert}{\sqrt{d}} = \frac{\sqrt{\sum_i \left(b_1^i - b_2^i \right)^2}}{\sqrt{d}}.
$$
For matrices with sizes $n \times m$, we use:
$$
\Delta_w = \frac{\lVert w_1 - w_2 \rVert}{\sqrt{n \cdot m}} = \frac{\sqrt{\sum_{i}^n \sum_j^m \left(w_1^{ij} - w_2^{ij}\right)^2}}{\sqrt{n \cdot m}}.
$$
Results showed in \Cref{fig:delta_w}. Note that two comparisons are plotted on the same scale. While the differences between the (a) pair are small, the differences between (b) can be observed.

The second experiment interpolates between two language models: DistilGPT-2 and GPT-C4. See \Cref{fig:distil_to_c4} for the results. We found that models obtained at every interpolation step completely forget the knowledge obtained during the training procedure. We additionally estimate the fraction of the distinct n-grams. At every point where perplexity becomes lower than initial values (0 and 1), we observe a significant drop in unique n-grams. The grammar score has two major peaks at points 0 and 1.

Models obtained by interpolating between different pre-trained models were found to fail at the basic language model tasks. This experiment confirms the importance of initializing fine-tuned models in the same way.

\section{Conclusion}

In our paper, we looked into simple linear weight interpolation between pre-trained and fine-tuned models, and concluded that this method performs surprisingly well. We found that different types of interpolation have different strengths and flaws, which we discuss in detail in the Experiments section. We have researched this phenomenon and provided intuition on why large language models, highly non-linear complex functions, are capable of generating texts with good metrics even after simple linear interpolation.

\bibliography{anthology,custom}
\bibliographystyle{acl_natbib}

\appendix

\section{Experiment Details}
\label{sec:appendix}

\subsection{Details of Controllable Text Generation Experiments} \label{sec:det_controllable_text_generation}

We fine-tuned two GPT-2 Large models on the SST dataset and ran a hyperparameter search using the grid from Table \ref{hyp-range-ft}.

\begin{table}[htb!]
\centering
\begin{tabular}{c|c} 
\toprule
Parameter                  & Values range                \\ 
\midrule
Learning rate              & {[}1e-4, 1e-5, 1e-6]  \\ 
\midrule
Batch size                 & {[}32, 64, 128, 256]             \\ 
\midrule
Steps                 & {[}500, 1000, 2000]       \\ 
\bottomrule
\end{tabular}
\caption{Hyperparameter search ranges used in fine-tuning.}
  \label{hyp-range-ft}
\end{table}

After the training, we proceeded with the best model in terms of perplexity on the corresponding validation sets. The best parameters are reported in Table \ref{best-hyp-ft}.

\begin{table}[htb!]
\centering
\begin{tabular}{c|cc} 
\toprule
Parameter                  & Positive ($\theta_+$)    & Negative ($\theta_-$)  \\ 
\midrule
Learning rate              & 1e-6  & 1e-6     \\ 
\midrule
Batch size                 & 64    & 64        \\ 
\midrule
Steps                 & 1000    & 1000       \\ 
\bottomrule
\end{tabular}
\caption{Best hyperparameters.}
  \label{best-hyp-ft}
\end{table}

As a Positive Text Score metric, we use outputs of the RoBERTa-base model trained by CardiffNLP\footnote{https://huggingface.co/cardiffnlp/twitter-roberta-base-sentiment} \cite{rosenthal2017semeval}. The model outputs consist of three probabilities: negative, neutral and positive sentiment. For the final score, we use the expectation of positive sentiment (see Equation \ref{pts}).

\begin{equation} \label{pts}
    score = 0 \cdot P(neg) + 0.5 \cdot P(neutral) + 1 \cdot P(pos)
\end{equation}

For perplexity, we use the GPT-2 XL model and count the perplexity of all generated texts. 

We also evaluate the Grammar Score using the RoBERTa-base model fine-tuned on the CoLA dataset by TextAttack\footnote{https://huggingface.co/textattack/roberta-base-CoLA} \cite{textattack}. The final score is the mean probability of the text being grammatically correct.

Text generation parameters can be found in Table \ref{gen_params}:
 
\begin{table}[htb!]
\centering
\begin{tabular}{c|c} 
\toprule
Parameter                  & Value              \\ 
\midrule
top-p              & 0.9  \\ 
\midrule
max new tokens                 & 30       \\ 
\bottomrule
\end{tabular}
\caption{Parameters used for text generation.}
  \label{gen_params}
\end{table}
\subsection{LM Training Details}\label{sec:lm_training_details}

\begin{figure}[h!]
  \centering
    \includegraphics[width=\linewidth]{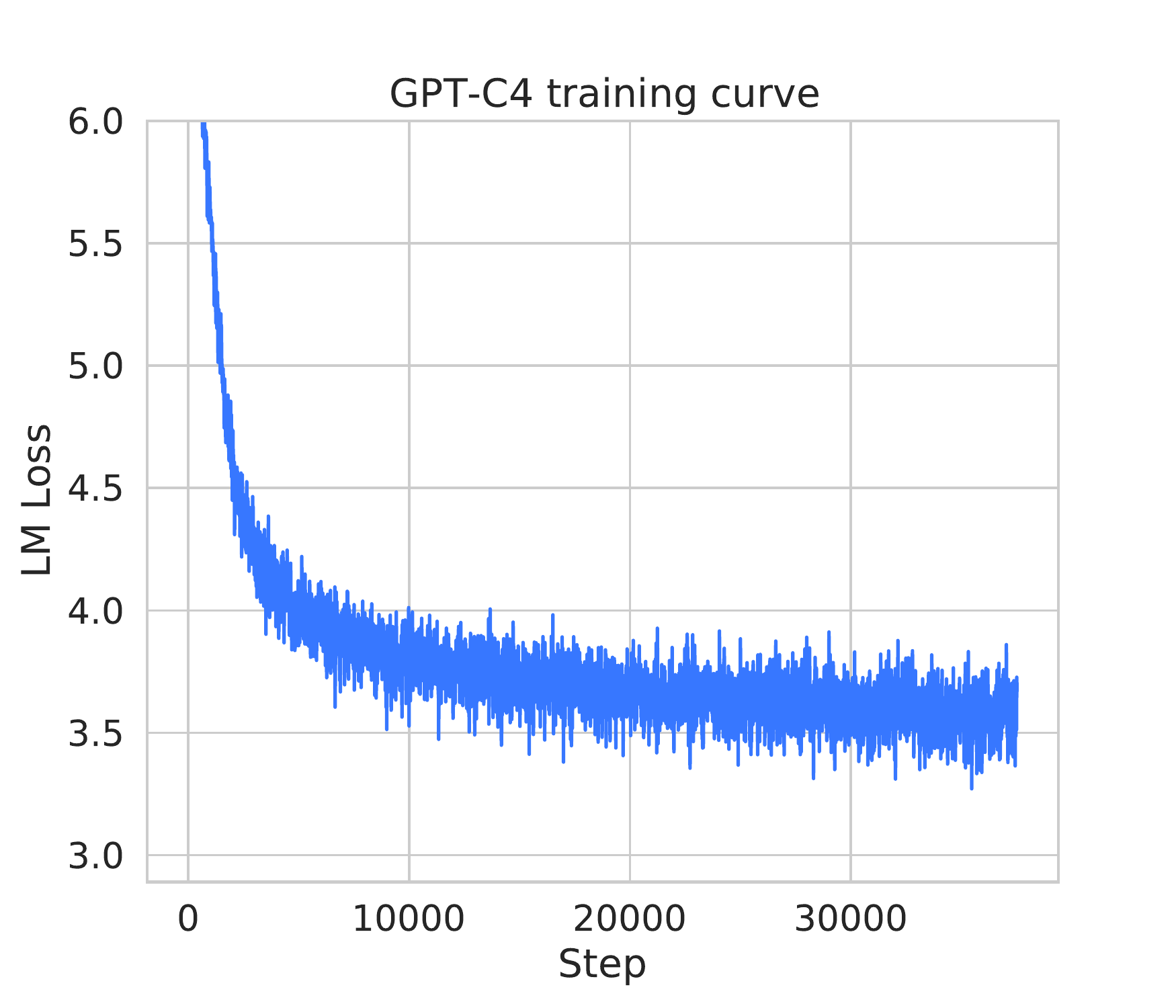}
    \caption{Loss dynamic during LM training.}
    \label{lm_loss}
 \end{figure}

We trained a GPT-Like language model on the C4 \cite{2019t5} dataset. This model's architecture is identical to the DistilGPT-2 model \cite{sanh2019distilbert}. We used 8x NVidia A100-SXM-80GB GPUs with bf16 mixed precision \cite{bf16}. We trained our model for 37K steps with the AdamW \cite{adamw} optimizer and a cosine scheduler with warmup. Parameters for the training procedure can be found in the table \ref{lm_params}. Model was trained until convergence, and the loss dynamic can be found in Figure \ref{lm_loss}.

\begin{table}[htb!]
\centering
\begin{tabular}{c|c} 
\toprule
Parameter                  & Value              \\ 
\midrule
Max LR              & 3e-4  \\
\midrule
Weight decay           & 0.01  \\
\midrule
$\beta_1$             & 0.9  \\
\midrule
$\beta_2$              & 0.95  \\
\midrule
$\varepsilon$           & 1e-8  \\ 
\midrule
Warmup steps           & 5000  \\
\midrule
Effective batch size                 & 1024       \\ 
\bottomrule
\end{tabular}
\caption{Parameters used for LM training.}
  \label{lm_params}
\end{table}

\end{document}